\documentclass[10pt,twocolumn,letterpaper]{article}

\usepackage{iccv}
\usepackage{times}
\usepackage{epsfig}
\usepackage{graphicx}
\usepackage{amsmath}
\usepackage{amssymb}
\usepackage{etoc}
\usepackage[accsupp]{axessibility}  

\usepackage{csquotes}
\usepackage{subfig}
\usepackage{threeparttable} 
\usepackage{multirow,tabularx} 
\usepackage{makecell}
\usepackage[normalem]{ulem}
\usepackage{color}
\usepackage{xcolor}

\usepackage[breaklinks=true,bookmarks=false]{hyperref}

\iccvfinalcopy 

\def\iccvPaperID{10604} 

\ificcvfinal\fi

\begin{document}

\title{View Consistent Purification for Accurate Cross-View Localization}

\author{Shan Wang$^{1,2}$ \quad Yanhao Zhang$^{1}$ \quad Akhil Perincherry$^{3}$ \quad Ankit Vora$^{3}$ \quad  Hongdong Li$^{1}$\\
{\tt\small $^{1}$Australian National University} \quad  {\tt\small$^{2}$Data61, CSIRO} \quad {\tt\small$^{3}$Ford Motor Company}
}

\maketitle
\ificcvfinal\thispagestyle{empty}\fi

\begin{abstract}
This paper proposes a fine-grained self-localization method for outdoor robotics that utilizes a flexible number of onboard cameras and readily accessible satellite images. 
The proposed method addresses limitations in existing cross-view localization methods that struggle to handle noise sources such as moving objects and seasonal variations. It is the first sparse visual-only method that enhances perception in dynamic environments by detecting view-consistent key points and their corresponding deep features from ground and satellite views, while removing off-the-ground objects and establishing homography transformation between the two views.
Moreover, the proposed method incorporates a spatial embedding approach that leverages camera intrinsic and extrinsic information to reduce the ambiguity of purely visual matching, leading to improved feature matching and overall pose estimation accuracy.
The method exhibits strong generalization and is robust to environmental changes, requiring only geo-poses as ground truth.
Extensive experiments on the KITTI and Ford Multi-AV Seasonal datasets demonstrate that our proposed method outperforms existing state-of-the-art methods, achieving median spatial accuracy errors below $0.5$ meters along the lateral and longitudinal directions, and a median orientation accuracy error below $2^\circ$
\footnote{Our project page is \href{https://shanwang-shan.github.io/PureACL-website/}{https://shanwang-shan.github.io/PureACL-website/}}.
\end{abstract}

\etocdepthtag.toc{mtchapter}
\etocsettagdepth{mtchapter}{subsection}
\etocsettagdepth{mtappendix}{none}

\section{Introduction}
Accurate self-localization is a fundamental problem in mobile robotics, particularly in the context of autonomous driving. 
While Global Positioning System (GPS) is a widely adopted solution, its accuracy hardly meets the stringent requirements of autonomous driving \cite{Reid_2019}.
Real-Time Kinematic (RTK) positioning systems provide an alternative by correcting GPS errors, but their implementation is hindered by the need for signal reference stations \cite{langley1998rtk}, rendering them an expensive solution.
On the other hand, odometry \cite{nister2004visual,engel2017direct,zhang2014loam,voraAerial} or simultaneous localization and mapping (SLAM) \cite{mur2017orb,kerl2013dense,shin2020dvl,voraAerial} methods can generate accurate short-term trajectories, however, they experience drift accumulation over time that can only be alleviated through loop closures if the agent's trajectories overlap.
Lastly, other self-localization techniques \cite{6942558,liu2017efficient,von2020lm,sarlin21pixloc} that rely on a pre-constructed 3D High Definition (HD) maps face limitations in terms of the extensive time and resources required for map acquisition and maintenance.

\begin{figure}
    \centering
    \begin{minipage}[t]{0.46\textwidth}
    \subfloat[\it Query]{
    \includegraphics[width=0.64\linewidth]{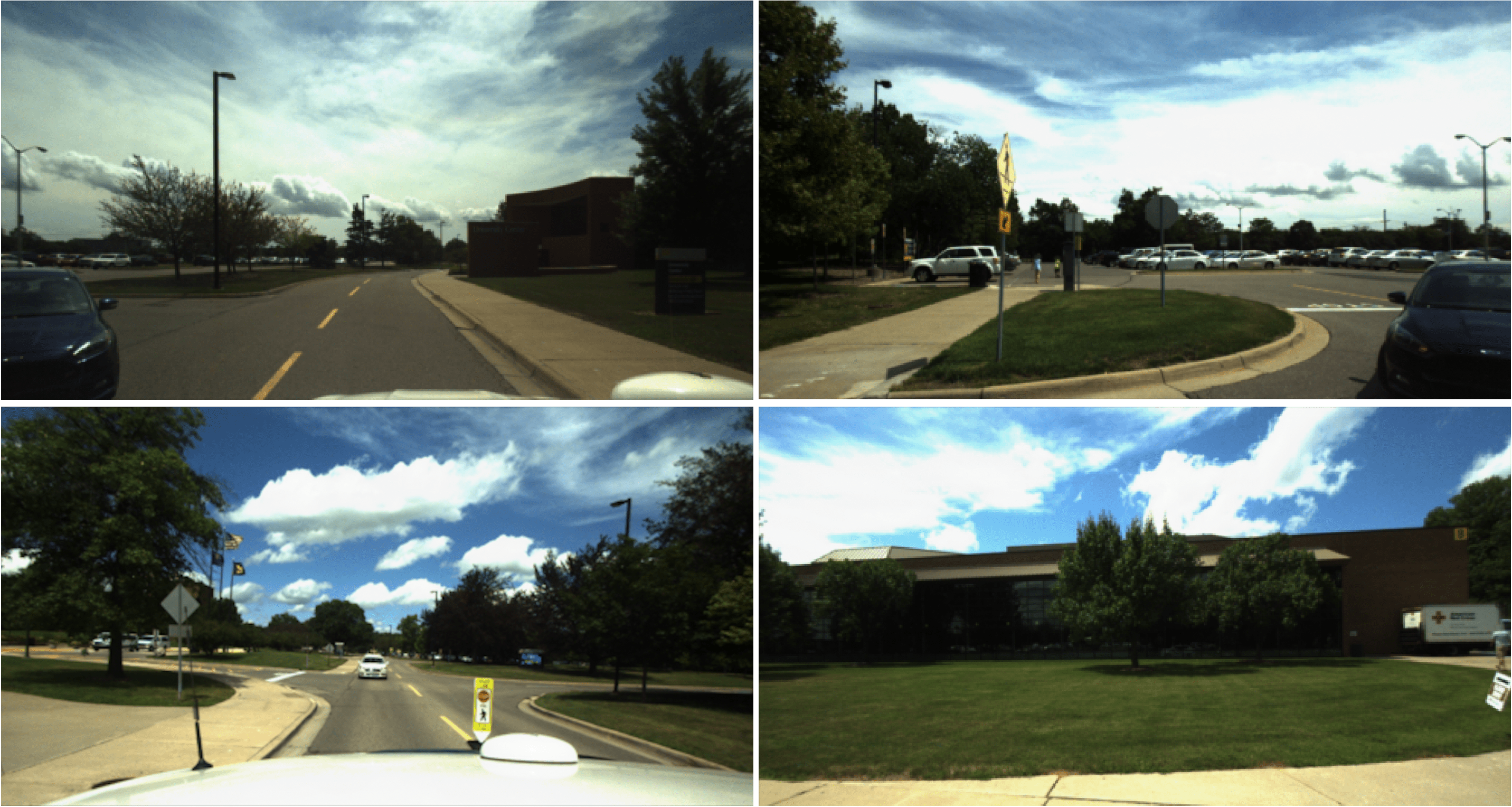}
    }
    \subfloat[\it Reference]{
    \includegraphics[width=0.345\linewidth]{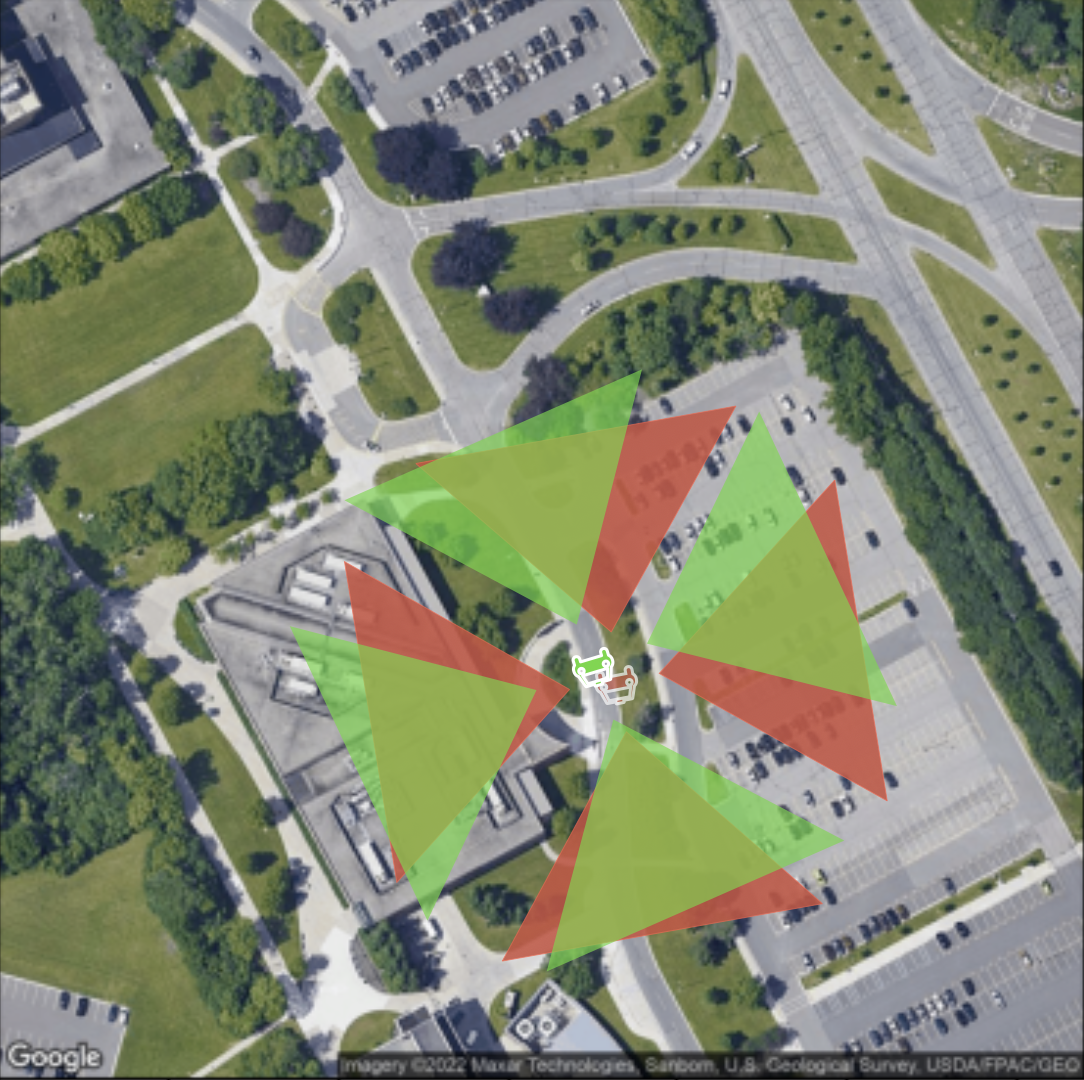}
    }
    \end{minipage}
    \caption{(a) Query ground view (onboard camera) images (front, rear, left, and right). (b) reference satellite image.
    The initial and ground truth poses, and FoV of cameras are shown in \textcolor{red}{(red)} and \textcolor{green}{(green)}, respectively.}
    \label{fig:Task}
    \vspace{-0.5cm}
\end{figure}

Using off-the-shelf satellite images as ready-to-use maps to achieve cross-view localization brings an alternative and promising way for low-cost localization.
However, due to the significant disparity between overhead views captured by satellites and views seen by robots, cross-view localization is more challenging than traditional methods.  
To address this, it is crucial to purify view-consistent features that can support the localization process. 
Furthermore, satellite views can be captured at different times, leading to variations in seasonal and temporal conditions. The cross-view consistent purification can also minimize the impact of moving and seasonal objects. 

Most previous cross-view localization methods \cite{shi2020optimal,hu2018cvm,Liu_2019_CVPR,toker2021coming,shi2020looking, zhu2021vigor} 
approach the task as an image retrieval problem, leading to coarse localization accuracy that is inferior to commercial GPS which can achieve an error of up to 4.9 meters
in open sky conditions \cite{van2015world}. In contrast, our method utilizes a coarse pose that is easily obtainable from the Autonomous Vehicles system,
to estimate the fine-grained 3-DoF (lateral, longitudinal, yaw) pose of the robot. This is accomplished through visual cross-view matching, utilizing ground-view images captured by onboard cameras and a spatially-consistent satellite map. Additionally, our method supports multiple camera inputs, which extend the field of view of the query robot. The setting is illustrated in Fig.~\ref{fig:Task}.

Our fine-grained visual localization method utilizes sparse (keypoint) feature matching, a departure from prior methods that rely on dense feature matching. To reduce the inherent ambiguity in purely visual matching, the method incorporates a camera intrinsic and extrinsic aware spatial embedding.
Homography transformation is used to establish correspondences between the two views.
An on-ground confidence map is employed to ensure the validity of the transformation and eliminate off-the-ground objects.
Additionally, a view consistency confidence map is utilized to mitigate the impact of moving objects and viewpoint variation.
The localization process begins with the extraction of spatially aware deep features and the generation of view-consistent, on-ground confidence maps for both views. View-consistent key points are then detected from the ground view confidence map and matched with their corresponding points in the satellite view. The optimal pose is determined through an iterative search using a differentiable Levenberg-Marquardt (LM) algorithm.

Using Google Maps \cite{google} as the satellite view, we evaluate our method on two datasets: the Ford Multi-AV Seasonal (FMAVS) \cite{Agarwal_2020} and the KITTI Datasets \cite{geiger2013vision}. The results demonstrate the superiority of our proposed method, achieving mean localization error of less than $\{ 0.14m, 3.57^\circ\}$ on KITTI with one front-facing onboard camera, and less than $\{ 0.88m, 0.74^\circ\}$ on FMAVS with four surrounding onboard cameras.

We summarize our contributions as below:
\begin{itemize}
    \item 
    the first sparse visual-only cross-view localization method that estimates accurate pose with low spatial and angular errors. 
    

    \item a view-consistent on-ground key point detector that reduces the impact of dynamic objects and viewpoint variations, as well as removes off-the-ground objects.

    \item a spatial embedding that fully utilizes camera intrinsic and extrinsic information to improve the extraction of spatially aware visual features.

    \item a multi-camera fusion approach that significantly improves localization accuracy.
\end{itemize}

\section{Related work}
\noindent\textbf{Depth Aware Accurate Cross-view Localization}.
The task of accurate cross-view localization has gained attention in recent years. 
Researchers have mainly focused on developing solutions for Radar and LiDAR cross-view localization as depth information helps in aligning the ground and satellite perspectives.
RSL-Net \cite{tang2020rsl} estimates the robot pose by registering Radar scans on a satellite image. This method was later extended to a self-supervised learning framework in \cite{tang2021self}.
Another work \cite{tang2021get} matches the top-down representation of a LiDAR scan with 2D points detected from satellite images. 
These methods have limitations and are only effective in environments with strong prior structure knowledge, failing in general, non-urban environments.
\cite{barsan2020learning} performs localization on bird's eye view (BEV) LiDAR intensity maps using deep feature matching between LiDAR scan and the intensity map. \cite{wei2019learning} extends this method by incorporating compressed binary maps.
Hybrid sensor solutions have also been explored, such as in \cite{miller2021any} where an aerial robot achieves global localization through the use of egocentric 3D semantically labelled LiDAR, IMU, and visual information. CSLA \cite{fervers2022continuous} and SIBCL \cite{wang2023satellite} extract visual features from ground and satellite images and use LiDAR points to establish correspondence between the two views.
CSLA \cite{fervers2022continuous} aims to estimate 2-DoF translation,
while SIBCL \cite{wang2023satellite} aims to estimate 3-DoF pose, including an additional orientation.
All these methods critically rely on depth information to build the correspondence across the two views. In contrast, our method is a visual-only solution that aims to achieve comparable localization accuracy using cheaper commodity sensors.

\noindent\textbf{Visual Accurate Cross-view Localization}.
Most visual-only cross-view localization methods rely on homography transformations of the ground plane, as they lack reliable depth information.
\cite{xia2022visual} aims to estimate 2-DoF translation using similarity matching and produces a dense spatial distribution to address localization ambiguities. HighlyAccurate \cite{shi2020beyond}  projects satellite features into the ground view and optimizes the robot pose through dense feature matching. 
One of its drawbacks is the limited ability to effectively eliminate outliers, such as noise caused by off-the-ground objects (which violates the assumption of homography transformation of the ground plane) and dynamic objects.
As a result, their overall performance is limited. In contrast, our method constructs geometric correspondences across sparse view-consistent on-ground keypoints, ensuring that the pose estimation is based on accurate correspondences leading to improved precision.

\section{Our Method} 

\begin{figure*}[!htb]
    \centering
    \includegraphics[width=0.95\linewidth]{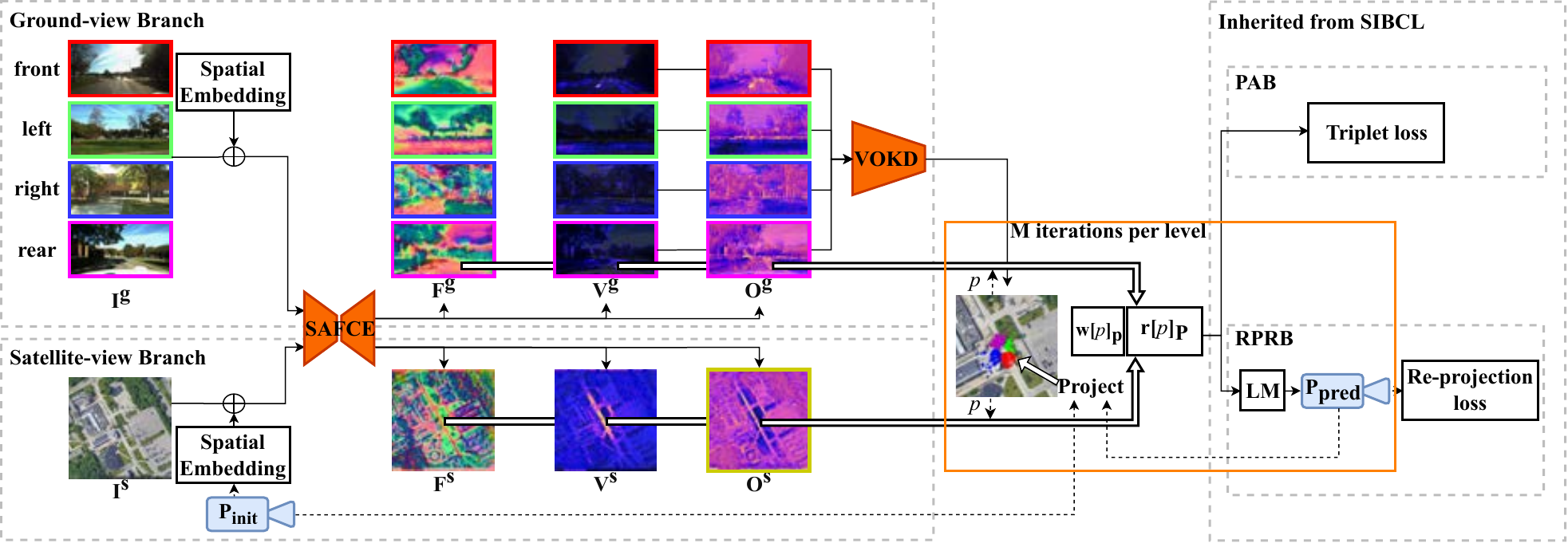}
    \caption{Overview of PureACL. SAFCE is used to produce feature maps ($F$), view-consistent confidence maps ($V$), and on-ground confidence maps ($O$) separately for satellite and ground-view images. 
    The VOKD fuses the confidence maps and identifies the top-k confident features from the ground-view images and their corresponding features on the satellite feature maps.
    Sub-pixel interpolation is used to lookup point features ($F[p]$ from $F$) and their weights ($w[p]$ from $V \otimes O$).
    The residual between the two views ($r[p]_{\mathbf{P}} = F^{s}[p]_{\mathbf{P}} - F^{g}[p]$) and the point weights ($w[p]_{\mathbf{P}} = w^{s}[p]_{\mathbf{P}} \times w^{g}[p]$) are fed to the RPRB for subsequent pose optimization. The \textcolor{olive}{olive} outline indicates that the $O^s$ disables gradient backpropagation while \textcolor{red}{red}, \textcolor{green}{green}, \textcolor{blue}{blue} and \textcolor{magenta}{magenta} outlines and points represent the \textcolor{red}{front}, \textcolor{green}{left}, \textcolor{blue}{right} and \textcolor{magenta}{rear} views, respectively.
    }
    \label{fig:architecture}
\end{figure*}

Our work aims to achieve fine-grained cross-view localization by accurately estimating the 3-DoF pose, denoted by $\mathbf{P}_{pred} = \{ \phi_{pred}, \varphi_{pred}, \theta_{pred} \}$, where $\phi$ and $\varphi$ represent lateral and longitudinal translations, respectively, and $\theta$ is the yaw angle. We are given a coarse initial pose $\mathbf{P}_{init} = \{ \phi_{init}, \varphi_{init}, \theta_{init} \}$, a reference satellite view image $I^{s}$, and a set of ground-view images $I^{g} = \{I^{i}\}_{i=1}^{N}$ captured by onboard cameras, where $N$ is the total number of onboard cameras \footnote{
Our method supports varying onboard camera quantities. In the experiments, we employed $N = 4$ for FMAVS and $N = 1$ for Kitti-CVL.
}. 
An overview of the proposed PureACL is shown in Fig.~\ref{fig:architecture}. It builds upon three innovative modules: 1) Spatially Aware Feature and Confidence Extractor (SAFCE) (Sec.~\ref{sec:SAFCE}), 2) View-consistent On-ground Keypoint Detector (VOKD) (Sec.~\ref{sec:VOKD}), and 3) Multi-camera Fusion (Sec.~\ref{sec:MQF}). Additionally, our approach utilizes two branches of objective functions inherited from the SIBCL method \cite{wang2023satellite}: the Pose-Aware Branch (PAB) and the Recursive Pose Refine Branch (RPRB). In the following sections, we provide a detailed explanation of each module. 

\subsection{Preliminary}
For completeness, we provide a brief description of the inherited PAB and RPRB.
The PAB utilizes a triplet loss \cite{qian2019softtriple} that encourages accurate pose (ground truth) and penalizes incorrect (initial) poses by differentiating the  residual between the ground truth and initial pose. Specifically, we compute the loss as follows:
\begin{equation}
L_{triplet} = log(1+e^{\alpha(1-\frac{\sum_{p}w[p]_{\mathbf{P}_{init}}\rho(\|r[p]_{\mathbf{P}_{init}}\|_2^2)}{\sum_{p}w[p]_{\mathbf{P}_{gt}}\rho(\|r[p]_{\mathbf{P}_{gt}}\|_2^2)})}),
\label{equ:triplet}
\end{equation}
where $\alpha$ is a hyper-parameter set to 10 based on experimental results, $\sum_{p}$ represents the sum of all key points, and $\rho$ is a robust cost function as defined in \cite{hampel2011robust}. 

The RPRB, on the other hand, aims to refine the initial pose iteratively using the LM algorithm to approach the ground truth pose.
It starts with the coarsest level and uses features from each level successively, with each subsequent level initialized with the output of the previous level.
Specifically, we update the pose as follows:
\begin{equation}
\delta_{t+1} = \delta_{t} -(\mathbf{H}+\lambda\text{ diag} (\mathbf{H}))^{-1}\mathbf{J}^\top \mathbf{W}\Upsilon,
\label{equ:delta}
\end{equation}
where $\delta$ represents an individual element in the 3-DoF pose. $t \in \{1, \cdots, M \times L\}$ represents the current iteration, and $M$ and $L$ represent the iteration count per level and the total number of levels, respectively.
The matrices $\Upsilon$ and $\mathbf{W}$ are formed by stacking the residuals  $r[p]_{\mathbf{P}}$ and weights $w[p]_{\mathbf{P}}$, while $\lambda$ is the  damping factors \cite{sarlin21pixloc}. The Jacobian and Hessian matrices are defined as follows: 
\begin{equation}
\mathbf{J} = \frac{\partial r[p]_{\mathbf{P}}}{\partial\delta}=\frac{\partial F^s[p]}{\partial[p^{s}_{2D}]_{\mathbf{P}}}\frac{\partial[p^{s}_{2D}]_{\mathbf{P}}}{\partial\delta} \text{ and }\mathbf{H}=\mathbf{J}^\top \mathbf{W}\mathbf{J},
\label{equ:Jacobian}
\end{equation}
where $[p^{s}_{2D}]_{\mathbf{P}}$ is the 2D projection of keypoints $p$ onto the satellite image using the pose $\mathbf{P}$, as shown in the right section of Fig.~\ref{fig:key_points}.
Finally, we supervise the optimized pose by computing the re-projection error as follows:
\begin{equation}
L_{reproject}(\mathbf{P}_{pred}) =  \sum\|[p^{s}_{2D}]_{\mathbf{P}_{pred}}-[p^{s}_{2D}]_{\mathbf{P}_{gt}} \|_2^2.
\label{equ:reprojection_loss}
\end{equation}

\subsection{Spatially Aware Feature/Confidence Extractor}\label{sec:SAFCE}

\begin{figure}
    \centering
    \includegraphics[width=0.49\textwidth]{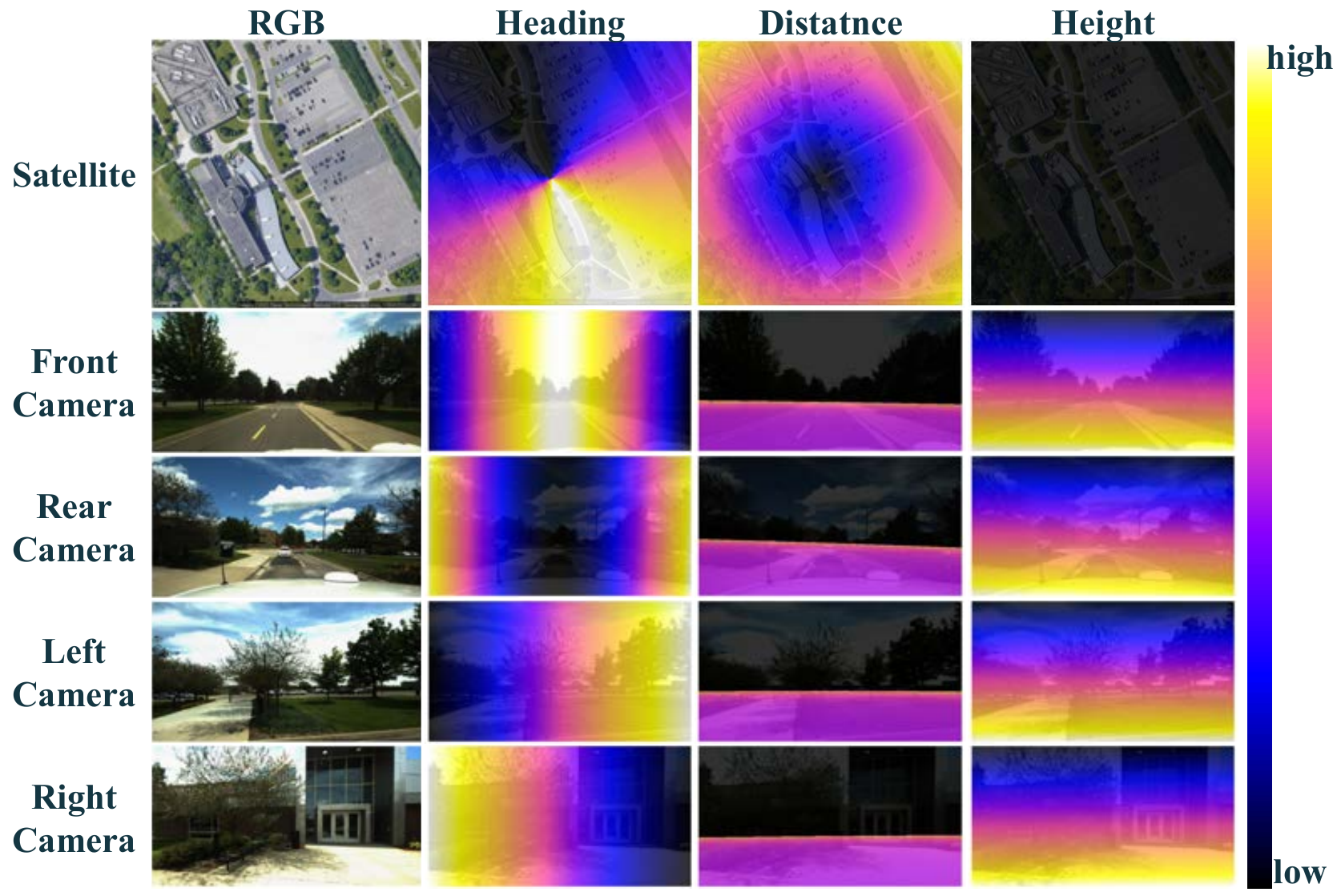}
    \caption{The Spatial Embedding. Heading is embedded using the cosine of its angle. Distance is embedded as the normalized on-ground distance from the robot, with the assumption that the pixel is lying on the ground. Height is normalized on the down axis of the ground view coordinates. For satellite images, the height is specifically set to a minimal value to indicate a top-down view.
     }
    \label{fig:embeding}
\end{figure}

Our approach improves the spatial embedding concept proposed in \cite{Liu_2019_CVPR} by leveraging the camera's intrinsic and extrinsic parameters to obtain highly accurate spatial information.
The spatial embedding $E^{g/s} \in \mathbb{R}^{h \times w \times 3}$ has $3$ channels: heading, distance, and height information.
The explanation of these channels is shown in Fig.~\ref{fig:embeding}. 
To incorporate additional spatial embedding information between the ground and satellite images, we transform the pixels in the onboard camera and satellite images into a common set of query world coordinates ( e.g., the GPS coordinates of the robot). In this coordinate system, the x-axis corresponds to the direction of motion, the y-axis points to the right, and the z-axis points downward. To perform this transformation, we use an inverse projection formula, which is shown in Eq.~\ref{equ:inv_intrinsic}:
\begin{equation}
    p_{3D}^{j2g} = \mathbf{R}_{j2g}\mathbf{K}_{j}^{-1}(p^{j}_{2D} \oplus 1),
    \label{equ:inv_intrinsic}
\end{equation}
where $\mathbf{K}_{j}$ is the intrinsic matrix of camera $j$, which can be either an onboard camera or a satellite camera $j \in \{i_{1}^{N},s\}$, and $\oplus 1$ concatenates 1 to generate the homogeneous coordinate.
The rotation from camera $j$ to the ground coordinate, $\mathbf{R}_{j2g}$, is obtained from the extrinsic information provided in the datasets for onboard  cameras and from the initial coarse pose for the satellite camera.
For onboard camera images, the 3D coordinate $p_{3D}^{i2g}$ is a homogeneous coordinate with an unknown scale, while for satellite images, 
$p_{3D}^{s2g}$ represents a world coordinate with an unknown down axis. This is because satellite images are approximated as parallel projections, and the equation for the calculating $p_{2D}^{s}$ is given by:
\begin{equation}
    p_{2D}^{s} = \begin{pmatrix}
    1/\gamma&0&c_u\\
    0&1/\gamma&c_v \\
    \end{pmatrix} p_{3D}^{s},
    \label{equ:s_intrinsic}
\end{equation} 
where $(c_u, c_v)$ represents the center of the satellite image, and $\gamma$ 
represents the meter-per-pixel ratio calculated using: 
\begin{equation}
    \gamma = \tilde{r}_{\text{earth}} \times \frac{\cos(\tilde{L} \times \frac{\pi}{180^{\circ}})}{2^{\tilde{z}} \times \tilde{s}},
    \label{equ:metre_per_pixel}
\end{equation}
where $\tilde{r}_{\text{earth}} = 156543.03392$ is radius of the Earth, $\tilde{L}$ is the latitude, $\tilde{z} = 18$ and $\tilde{s} = 2$ is the zoom factor and the scale of Google Maps \cite{google}, respectively.

\begin{figure}[!htb]
    \centering
        \includegraphics[width=0.50\textwidth]{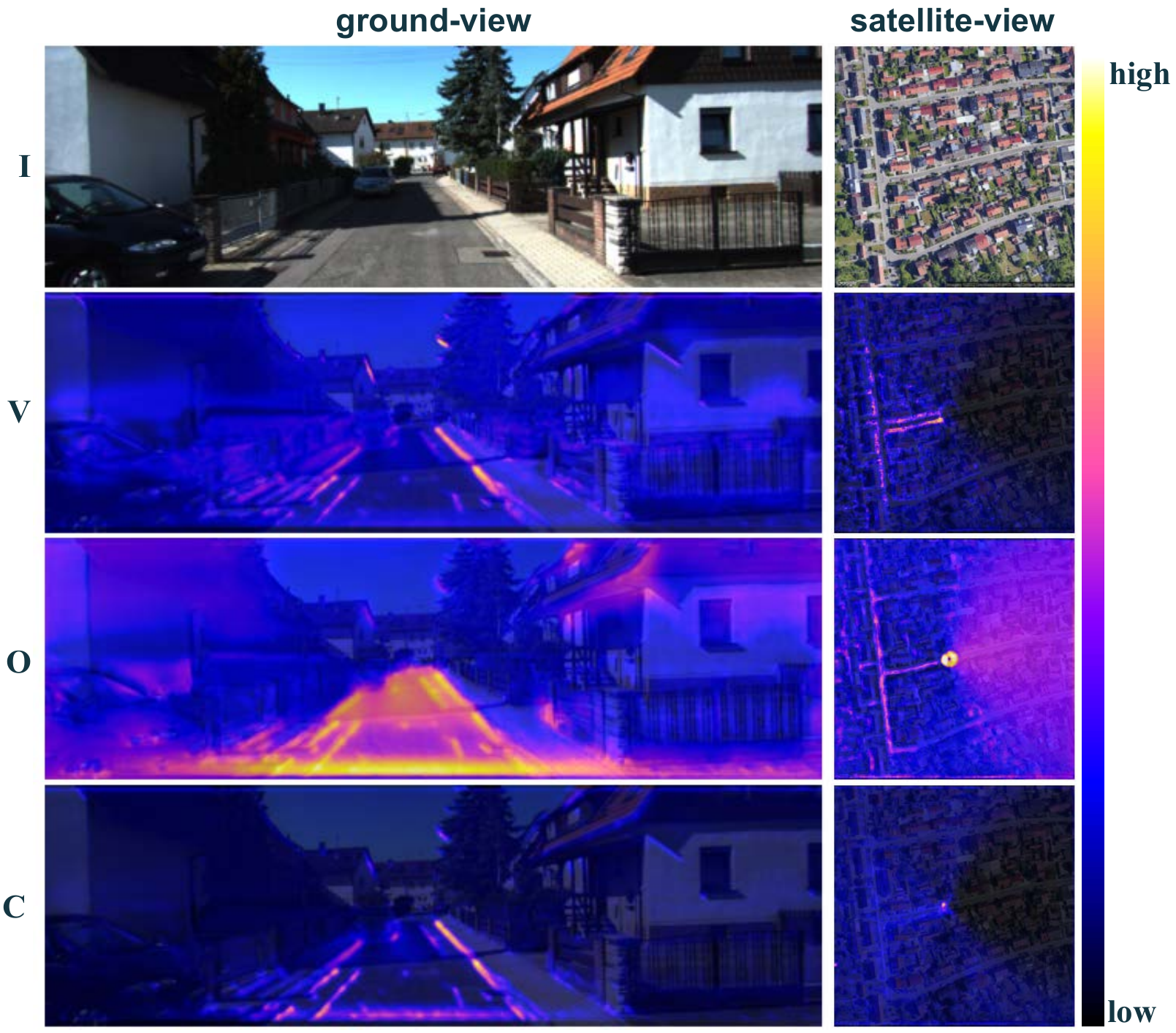}
    \caption{Illustration of confidence maps. The view-consistent confidence map ($2^{nd}$ row) $V$ assigns high confidence to objects that appear consistently in both ground-view and satellite images, such as road marks, curbs, and building roofs. Conversely, the confidence map assigns low confidence to temporally inconsistent objects, such as vehicles. The on-ground confidence map ($3^{rd}$ row) $O$ highlights only on-ground cues, such as road marks and curbs. It is noteworthy that the area behind the robot is assigned a high score due to a lack of supervision, but it does not affect localization accuracy. This is because the influence of the area is suppressed by the view-consistent confidence ($w[p]$ from $V \times O$). The fused confidence map ($4^{th}$ row) $C$ highlights objects that are both view-consistent and on-ground. 
    }
    \label{fig:confidence}
    \vspace{-0.3cm}
\end{figure}

The heading information is embedded using the cosine value, which is symmetric to both positive and negative orientation noise. This enables distinction between 360-degree views, calculated using the x-axis ($p_{3D}^{j2g}[0]$) and  y-axis ($p_{3D}^{j2g}[1]$) through trigonometric functions, as shown below:
\begin{equation}
    E^{j}[0] =  p_{3D}^{j2g}[0] / \sqrt{p_{3D}^{j2g}[0]^2 + p_{3D}^{j2g}[1]^2}.
    \label{equ:heading_E}
\end{equation}
The normalized distance embedding of ground images is obtained by assuming all pixels lie on the ground plane:
\begin{equation}
    E^{j}[\text{1}] = \sqrt{p_{\tilde{3D}}^{j2g}[0]^2 + p_{\tilde{3D}}^{j2g}[1]^2} / \mathcal{D}, 
    \label{equ:distance_E}
\end{equation}
where $\mathcal{D}$ is the maximum visible distance, set to 200 meters according to the satellite maps size and 
\begin{equation}
    p_{\tilde{3D}}^{i2g} = \frac{h_i}{p_{3D}^{i2g}[2]} \times p_{3D}^{i2g} + \mathbf {t}_{i2g} \text{ and  } p_{\tilde{3D}}^{s2g} = p_{3D}^{s2g} + \mathbf{t}_{s2g},
    \label{equ:3d_scale}
\end{equation}
where $h_i$ is the onboard camera height relative to the ground plane. 
For ground view images, the height embedding $E[2]$ is equal to the value along the down axis, represented as $p_{3D}^{i2g}[2]$. In the case of satellite images, we set the height embedding to the minimal value to indicate a top-down perspective. Fig.~\ref{fig:confidence} demonstrates that our approach effectively directs greater attention towards the features located in front of the robot by leveraging spatial embedding when using only the front onboard camera.

The SAFCE employs a U-Net structure ($\mathcal{F}_{\nu}$) to extract the satellite and ground-view feature maps, represented as $F^{j} = \mathcal{F}_{\nu}(I^{j} \oplus E^{j})$, where $j \in \{i_{1}^{N},s\}$,
and $\oplus$ denotes channel concatenation.
The maps are then processed by a convolutional layer followed by a reverse sigmoid active function  ($\mathcal{C}_{\psi}$) to produce view-consistent confidence maps ($V^{j}$) and on-ground confidence maps ($O^{j}$) represented as $V^{j},O^{j} = \mathcal{C}_{\psi}(F^{j})$. 
Each map has multiple resolutions, for example, $F = \{F_{l} \in \mathbb{R}^{h_{l} \times w_{l} \times c_{l}}\}_{l=1}^{L}$ ($\mathbb{R}^{h_{l} \times w_{l}}$ for $V$ and $O$), where $L = 3$ is adopted in our setting. 
The maps are ordered from coarsest to finest level as $l = \{1, 2, 3\}$. The feature and confidence extraction from each image is performed in parallel using a shared-weight model, allowing for a flexible number of onboard cameras (N).


The view-consistent confidence map $V$ represents the confidence of objects appearing in both satellite and ground-view images. $V$ is used as a multiplying factor for the point weights supervised by PAB and RPRB, and is penalized through the network training for the points with high residual (indicating distinct features between the cross-view). Considering the temporal gap between the two views, $V$ effectively filters out objects that are temporally or seasonally inconsistent, e.g. vehicles, pedestrians, and leaves. Additionally, it highlights view consistent reference objects, including road marks, lanes, building edges, and tree roots. An example is shown in Fig.~\ref{fig:confidence} (row 2). More visualizations are shown in the supplementary. 

The on-ground confidence map $O$ is designed to validate the homography transformation between the ground and satellite views. As a multiplying factor for the point weights, off-ground points that cause incorrect Geo-correspondence between the ground and satellite views, resulting in high residuals, have their on-ground confidence penalized to reduce the overall loss.
Given that an incorrect height assumption in points can lead to erroneous projections on the satellite map, penalizing the satellite on-ground confidence map is not meaningful. So we only apply the backpropagation to the ground-view on-ground confidence map. 
An example of the learned confidence maps is shown in Fig.~\ref{fig:confidence} (row 3).

\subsection{View-consistent On-ground Keypoint Detector}\label{sec:VOKD}
\begin{figure}[!htb]
    \centering
    \includegraphics[width=0.49\textwidth]{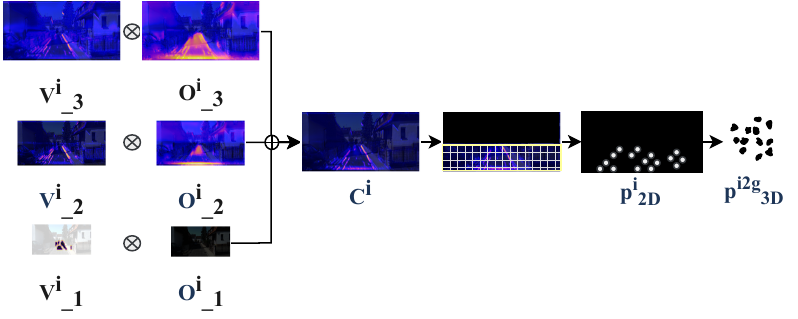}
    \caption{The pipeline of VOKD. It begins with confidence map fusion, in which all level confidence maps from the view-consistent and on-ground maps are combined to create a single map. Next, in the 2D keypoint detection step, the top part of the image is ignored to concentrate on the ground plane. Moreover, a max pooling technique is employed to avoid overly crowded keypoint detection. Finally, based on the assumption that all detected points are on the ground, their 3D ground query coordinates are calculated.
     }
    \label{fig:point_detector}
    \vspace{-0.2cm}
\end{figure}
Fig.~\ref{fig:point_detector} illustrates the details of the proposed VOKD. The view-consistent and on-ground confidence maps of different resolutions are fused to generate the final confidence:
%
\begin{equation}
    C^{i} = \sum_{l=1}^{L}{\Xi(\mathcal{N}(V^{i}_{l} \otimes O^{i}_{l}), (h_L,w_L))},
    \label{equ:C_merge}
\end{equation}
where $h_{L}$ and $w_{L}$ represents the resolution of the fine level confidence map, $\Xi$ is an interpolation function, and $\mathcal{N}$ is a min-max normalisation, and $\otimes$ represents element-wise multiplication.
The bottom row of Fig.~\ref{fig:confidence} demonstrates the efficacy of the fused confidence map in filtering out off-the-ground objects and emphasizing temporal stability and view consistency in cues such as road markings and curbs for the subsequent pose estimation. 
More visual examples can be found in the supplementary.

\begin{figure}[!htb]
    \centering
     \includegraphics[width=0.49\textwidth]{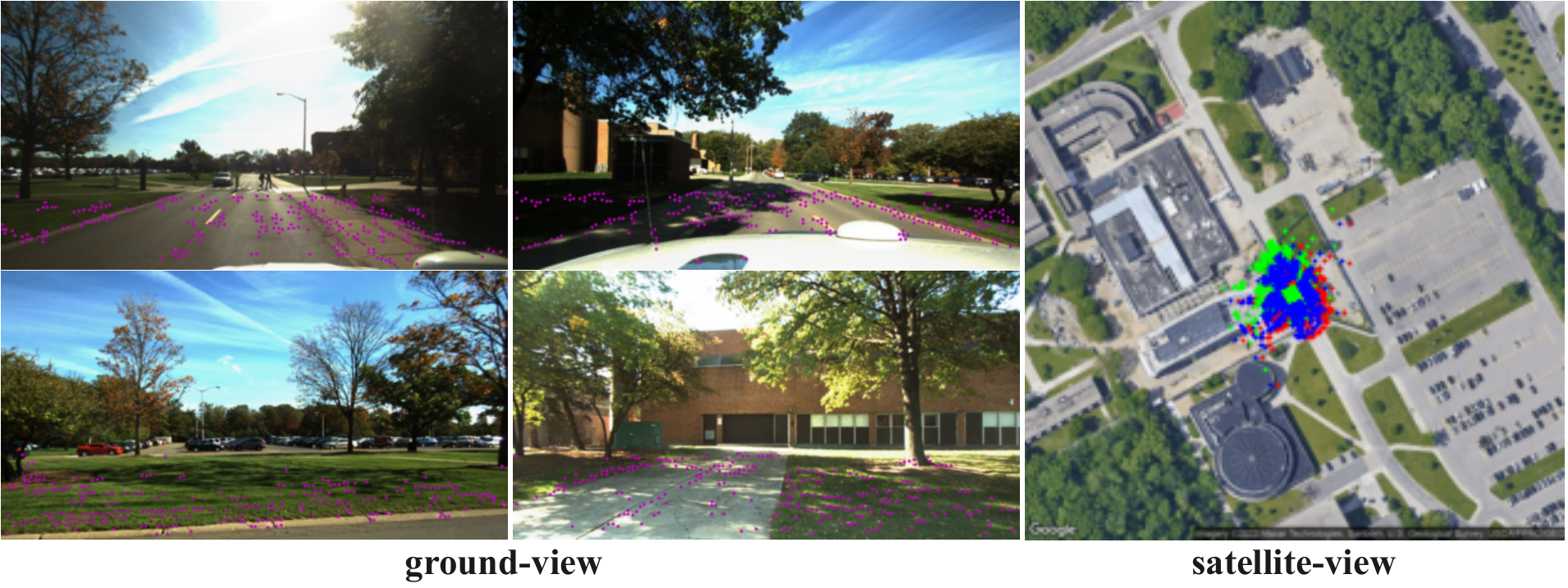}
    \caption{(left) On-ground keypoints on ground-view images and detected keypoints \textcolor{magenta}{(magenta)}. (right) Projection of on-ground keypoints on the satellite image.
    Projection by initial pose is shown in \textcolor{red}{(red)}, projection by predicted pose is shown in \textcolor{blue}{(blue)}, and projection on ground truth pose is shown in \textcolor{green}{(green)}. 
    }
    \label{fig:key_points}
    \vspace{-0.3cm}
\end{figure}

In order to achieve on-ground keypoint detection, our focus is limited to the area below the focal point, which corresponds to the on-ground area and is our primary interest. From this area, we select the top-K points with the highest confidence score from the fused confidence map. To avoid overcrowding of keypoints, we partition the fused confidence map into smaller patches of size $8\times8$ and enforce a limit of one detected keypoint per patch. This approach ensures that the selected keypoints are well-distributed across the on-ground area, thereby improving the accuracy of subsequent pose estimation.
The left part of Fig.~\ref{fig:key_points} displays the detected view-consistent on-ground 2D keypoints.
These 2D keypoint coordinates $p_{2D}^{i}$ are used to calculate their corresponding 3D ground world coordinates  $p_{\tilde{3D}}^{i2g}$ through the equations Eq.~\eqref{equ:inv_intrinsic} and Eq.~\eqref{equ:3d_scale}. 
The right part of Fig.~\ref{fig:key_points} shows the projection of these 3D coordinates onto the satellite image ($p_{2D}^{s} = \mathbf{K}_{s}(\mathbf{R}_{g2s}p_{\tilde{3D}}^{i2g}+\mathbf{t}_{g2s}$)).

\subsection{Multi-camera Fusion}\label{sec:MQF}
Our method is flexible and can handle multiple
cameras as input, without any restrictions on the field of view. In case there is a potential overlap between the views captured by adjacent cameras, keypoints detected in one camera may be visible in another camera as well. In such cases, we select the point feature with the highest weight:
\begin{equation}
    w^{g}[p] = \max_{i}^{N}(V^{i} \otimes O^{i})[p^{i}_{2D}],
    \label{equ:query_select_c}
    \vspace{-0.6cm}
\end{equation}
\begin{equation}
    F^{g}[p] = F^{i}[p_{2D}^{i}], ~~i = \arg\max_{i}^{N}(V^{i} \otimes O^{i})[p^{i}_{2D}].
    \label{equ:query_select_c2}
\end{equation}

\section{Datasets}
To evaluate the effectiveness of the proposed method, we followed the existing methods \cite{shi2020beyond, wang2023satellite} and conducted experiments on two widely used autonomous driving datasets: the 
FMAVS dataset \cite{Agarwal_2020} and KITTI dataset \cite{geiger2013vision}. We adopted the augmentation method proposed by \cite{wang2023satellite}, which involved incorporating spatially-consistent satellite images obtained from Google Maps \cite{google} using the GPS tags provided in the datasets. The satellite images had a resolution of $1,280 \times 1,280$ pixels and a scale of 0.22m per pixel for FordAV-CVL, and 0.2m per pixel for KITTI-CVL.

\begin{table*}[!htb]
\renewcommand{\arraystretch}{1.0}
\caption{Comparison on the KITTI-CVL dataset} 
\vspace{-0.2cm}
\label{Tab:KITTI_Comparison}
\centering
\footnotesize 
\setlength\tabcolsep{1pt}
\begin{tabular*}{\textwidth}{c||@{\extracolsep{\fill}}c c c c c c|c c c c c c|c c c c c}
\hline
& \multicolumn{6}{c|}{\bfseries Lateral} & \multicolumn{6}{c|}{\bfseries Longitudinal} & \multicolumn{5}{c}{\bfseries Yaw} \\
& mean$\downarrow$ & median$\downarrow$ & 0.25m$\uparrow$ & 0.5m$\uparrow$ & 1m$\uparrow$ & 2m$\uparrow$ &  mean$\downarrow$ & median$\downarrow$ &0.25m$\uparrow$ & 0.5m$\uparrow$ & 1m$\uparrow$ & 2m$\uparrow$ & mean$\downarrow$ & median$\downarrow$ & $1^\circ\uparrow$ & $2^\circ\uparrow$ & $4^\circ\uparrow$ \\
\hline\hline
$\star$ SIBCL\cite{wang2023satellite} & 1.02 & 0.54  & 25.59 & 46.26 & 72.63 & 89.78 & 1.69 & 0.64  & 21.91 & 41.22 & 64.47 & 80.37 & \textbf{1.91} & \textbf{0.85} & \textbf{56.05} & \textbf{79.70} & \textbf{90.89}\\
CVML\cite{xia2022visual} & 3.38 & 2.40 & 6.11 & 12.24 & 23.78 & 44.14 & 3.54 & 2.46  & 5.97 & 11.68 & 23.73 & 43.36 & - & - & - & - & -\\
HighlyAcc\cite{shi2020beyond} & 1.24 & 0.83  & 16.51 & 32.05 & 57.65 & 83.11 & 2.44 & 2.01 & 7.14 & 14.11 & 27.41 & 49.94 & 3.23 & 1.82  &29.83 & 53.41 & 76.51\\
Ours & \textbf{0.14} & \textbf{0.12} & \textbf{84.58} & \textbf{99.54} & \textbf{99.98} & \textbf{100.00} & \textbf{0.10} & \textbf{0.09} & \textbf{98.55} & \textbf{100.00} & \textbf{100.00} & \textbf{100.00} & 3.57 & 1.78  & 31.18 & 54.13 & 76.00\\
\hline
\end{tabular*}
\begin{tablenotes}
\item[0] $\star$: indicates LiDAR-visual hybrid methods. $\uparrow$: larger is better. $\downarrow$: lower is better.
\item[1] Our method significantly improves translation accuracy while maintaining orientation accuracy compared to SOTA visual method \cite{shi2020beyond}.
\end{tablenotes}
\vspace{-0.2cm}
\end{table*}

\begin{table*}[!htb]
\renewcommand{\arraystretch}{1.0}
\caption{Comparison on the FordAV-CVL dataset} 
\vspace{-0.2cm}
\label{Tab:FordAV_Comparison}
\centering
\footnotesize 
\setlength\tabcolsep{1pt}
\begin{tabular*}{\textwidth}{c c||@{\extracolsep{\fill}}c c c c c c|c c c c c c|c c c c c}
\hline
& & \multicolumn{6}{c|}{\bfseries Lateral} & \multicolumn{6}{c|}{\bfseries Longitudinal} & \multicolumn{5}{c}{\bfseries Yaw} \\
& & mean$\downarrow$ & median$\downarrow$ & 0.25m$\uparrow$& 0.5m$\uparrow$ & 1m$\uparrow$ & 2m$\uparrow$ & mean$\downarrow$ & median$\downarrow$ & 0.25m$\uparrow$ & 0.5m$\uparrow$ & 1m$\uparrow$ & 2m$\uparrow$ & mean$\downarrow$ & median$\downarrow$ & $1^\circ\uparrow$ & $2^\circ\uparrow$ & $4^\circ\uparrow$ \\
\hline\hline
\multirow{7}{*}{\makecell[c]{\rotatebox{90}{\textbf{Log4}}}} 
 &$\star$ SIBCL\cite{wang2023satellite} & 1.29 & 0.55  & 24.83 & 45.90 & 74.06 & 89.14 & 2.31 & 0.78  & 18.72 & 34.11 & 58.26 & 75.44 & 2.23 & 0.57  & 66.76 & 81.78 & 90.50 \\
&CVML\cite{xia2022visual} & 2.78 & 2.22  & 5.91 & 11.78 & 23.27 & 45.06  & 3.24 & 2.66 & 6.07 & 11.45 & 21.22 & 38.82 & - & - & - & - & -\\
&HighlyAcc\cite{shi2020beyond} & 1.21 & 0.84  & 16.56 & 31.31 & 57.64 & 85.45  & 2.47  & 1.82 & 7.11 & 13.87 & 28.53 & 53.64 & 2.94 & 1.83  & 30.74 &  53.08 & 78.40\\
&Ours (Front) & 0.94 & 0.54  & 26.11 & 46.73 & 73.48 & 89.69 & 1.56 & 0.80  & 17.95 & 34.44 & 56.30 & 75.41 & 2.77 & 1.18  & 44.47 & 66.61 & 83.26 \\
&Ours (2FR) & 0.60 & 0.51  & 24.66 & 48.78 & 82.40 & 98.20 & 0.99 & 0.65  & 22.45 & 41.31 & 64.49 & 86.69 & 1.14 & 0.77  & 60.78 & 85.28 & 96.28 \\
&Ours (2Sides) & 0.78 & 0.55  & 24.75 & 46.36 & 75.01 & 94.91 & 1.58 & 0.92  & 14.68 & 28.77 & 52.68 & 73.52 & 3.56 & 2.14 & 24.94 & 47.39, & 71.50\\
&Ours (4Cams) & \textbf{0.58} & \textbf{0.46} & \textbf{26.45} & \textbf{53.60} & \textbf{85.00} & \textbf{98.81} & \textbf{0.88} & \textbf{0.49} & \textbf{25.77} & \textbf{50.31} & \textbf{75.44} & \textbf{91.53}  & \textbf{0.74} & \textbf{0.50}& \textbf{77.61} & \textbf{94.94} & \textbf{98.57}\\
\hline
\multirow{4}{*}{\makecell[c]{\rotatebox{90}{\textbf{Log4$\rightarrow$5}}}} 
&$\star$ SIBCL\cite{wang2023satellite}  & 1.99 & 1.38 & 10.49& 21.57 & 39.05 & 64.98 & 6.27 & 3.23  & 13.77 & 22.22 & 31.11 & 42.62 & 3.32 & 1.78  & 31.78 & 54.91 & 78.90 \\
&CVML\cite{xia2022visual} & 3.10 & 2.31  & 5.25 & 10.88 & 20.58 & 43.25  & 3.32 & 2.63 & 5.89 & 10.45 & 21.86 & 39.87 & - & - & - & - & -\\
& HighlyAcc\cite{shi2020beyond} & 1.69 & 1.61  & 9.45 & 18.03 & 31.72 & 66.06  & 2.99 & 2.32 & 4.63 & 9.69 & 19.89 & 39.28  & 3.35 & 2.44 & 22.43 & 42.32 & 75.19\\
& Ours (4Cams)  & \textbf{0.96} & \textbf{0.68} & \textbf{20.03} & \textbf{37.83} & \textbf{65.09} & \textbf{87.53} & \textbf{1.43} & \textbf{0.82} & \textbf{17.45} & \textbf{33.91} & \textbf{56.73} & \textbf{76.96} & \textbf{2.76} & \textbf{1.38} & \textbf{39.20} & \textbf{61.90} & \textbf{79.37}\\
\hline
\end{tabular*}
\begin{tablenotes}
\item[0] Log4: Localization on the same road with the different time and seasons w.r.t. the training dataset.
\item[1] Log4$\rightarrow$5: Localization on a totally different road w.r.t. the training dataset to evaluate the generalization ability.
\item[2] We evaluate the effect of camera configuration on localization accuracy using multiple camera settings in Log4 (row 4-7).
\end{tablenotes}
\vspace{-0.3cm}
\end{table*}

In the FMAVS dataset, 
we utilized query images from four cameras (front left, rear right, side left, and side right) to capture the surrounding environment, providing an almost 360-degree field of view with minimal overlap.
Since the KITTI dataset provides only front-facing stereo camera images, we used the images from the left camera of the stereo pair as query images.
The FMAVS
includes multiple vehicle traversals over a consistent route. To evaluate our proposed method, we split the three traversals of the `Log4' trajectory into training, validation, and test sets, following the split strategy described in \cite{wang2023satellite}.
The KITTI dataset \cite{geiger2013vision} comprises various trajectories taken at different times. To assess our model's generalization ability, we selected test sets from different trajectories based on \cite{shi2020beyond}.

\section{Experiments}
\noindent\textbf{Metrics}.
Our objective is to estimate the 3-DoF pose, which includes lateral, longitudinal, and yaw information. We measure the accuracy of our proposed method by reporting the median and mean errors in lateral and longitudinal translations (in meters) and yaw rotation (in degrees).
In addition to these metrics, we also follow the evaluation criteria outlined in \cite{wang2023satellite} and report the average localization recall \footnote{The percentage of the prediction pose that is within a certain range.} at distances of 0.25m, 0.5m, 1m, and 2m, as well as at yaw rotation angles of $1^\circ$, $2^\circ$, and $4^\circ$.

\noindent\textbf{Implementation Details}.
In our experiments, we use an input size of $432 \times 816$ for the ground-view images in the Multi-AV Seasonal Dataset, and $384 \times 1248$ for the KITTI Dataset.
RTK GPS \footnote{RTK GPS achieves an accuracy of $2$ cm or better \cite{feng2008gps}.} is used as the ground truth pose. We add some noise to the RTK GPS poses to generate the initial pose. 
Unless otherwise stated, the initial pose is randomly sampled with a yaw angle error of $\pm 15^\circ$ and lateral, longitudinal shifts of $\pm 5$ meters, as the accuracy of GPS is within $4.9$ meters in open sky conditions \cite{van2015world}.
We detect $256$ ground keypoints from each input ground-view image. We set the batch size to $b=3$ for training on an NVIDIA RTX 3090 GPU, and use the Adam optimizer \cite{kingma2014adam} with a learning rate of $10^{-4}$. The feature extractor weights are initialized with the pre-trained weights from \cite{wang2023satellite}, which are trained on the KITTI-CVL dataset. The weights of the confidence generator are initially randomly initialized to values near $0$. Through the application of the inverse sigmoid activation function, these weights are tuned to initialize the confidence values in proximity to $50\%$.

\noindent\textbf{Inference Speed}.
The SAFCE processes four query ground-view images and one satellite image in approximately $200$ms. The detection time for all ground keypoints is about $3.5$ms. The optimization process, which runs for $20$ iterations at each of the three levels, takes a total of approximately $200$ms.

\noindent\textbf{Qualitative Results}.
We compare our method with recent state-of-the-art (SOTA) visual-only methods, CVML \cite{xia2022visual} and HighlyAccurate \cite{shi2020beyond}, as well as the LiDAR-visual hybrid method SIBCL \cite{wang2023satellite}. 
We present the evaluation results on the KITTI-CVL and FordAV-CVL datasets in Tab.~\ref{Tab:KITTI_Comparison} and Tab.~\ref{Tab:FordAV_Comparison}. 
To ensure a fair comparison, we trained HighlyAccurate \cite{shi2020beyond} and SIBCL \cite{wang2023satellite} under the same image resolution and initial pose noise range. Since CVML \cite{xia2022visual} is unable to accurately estimate fine-grained orientations, we only evaluated its performance in terms of location estimation. We trained their model with ground truth orientation.

Tab.~\ref{Tab:KITTI_Comparison} presents an evaluation of our method's ability to generalize to previously unseen routes in the KITTI-CVL dataset using a front camera. For translation accuracy, our method exhibits superior performance compared to SOTA methods, with a significant reduction in the translation error. Specifically, our method achieves a reduction of $86\%$ and $94\%$ in  mean lateral and longitudinal localization error. 
While our orientation accuracy is slightly less accurate than the LiDAR-based method, it maintains a comparable performance to SOTA visual-only method \cite{shi2020beyond} in terms of rotation error.
These results demonstrate the ability of our method to generalize to a wide range of scenes.

The performance of our method on cross-season generalization is presented in `Log4'\footnote{The trajectory of `Log4' was selected for method evaluation in SIBCL \cite{wang2023satellite} due to its relatively good satellite view alignment. Additionally, we evaluated other logs and the evaluation results can be found in the supplementary material.} of Tab.~\ref{Tab:FordAV_Comparison}. The test set in this case includes data from different time and seasons compared to the training set, which allows us to evaluate the performance of our method under varying lighting and seasonal conditions. Furthermore, in `Log4$\rightarrow$5' of Tab.~\ref{Tab:FordAV_Comparison}, we analyze our method's generalization capability on an unseen route. In both cases, our method outperforms existing SOTA methods by significant margins. Specifically, we achieve a reduction of $52\%$ and $43\%$ in mean localization lateral error, $62\%$ and $52\%$ in mean localization longitudinal error, and $67\%$ and $17\%$ in mean orientation error in terms of seen and unseen routes, respectively. These results once again demonstrate the strong performance and robust generalization capabilities of our proposed method.

\noindent\textbf{Performance with Varying Numbers of Camera Inputs}.
We investigate the impact of multiple onboard cameras on the FordAV-CVL dataset and evaluate our method using different camera setups. These setups include the front camera (Front) in the 1-camera setting, two side cameras (2Sides), the front and rear cameras (2FR) in the 2-camera setting, and all front, rear, and two side cameras (4Cams) in the 4-camera setting. 
Our findings indicate that even with the use of a single front camera (`Ours (Front)' in Tab.~\ref{Tab:FordAV_Comparison}), our method outperforms the SOTA methods. 
Additional camera inputs lead to further improvements in performance, particularly with regards to orientation estimation, which can be attributed to the fact that a larger field of view (FoV) provides more information to accurately estimate orientation.
Furthermore, our study reveals that the front and rear cameras provide more information for localization, whereas the left and right cameras contribute more to the lateral estimation. This could be attributed to the limited visibility of noticeable localization features such as road marks in the side cameras or the sensitivity of the side cameras to the roll angle.
It is noteworthy that our method, despite utilizing four onboard cameras, consumes less memory (4499 MB) than HighlyAccurate \cite{shi2020beyond}, which requires 6445 MB due to its use of sparse purification.

\begin{figure}[!htb]
    \centering
    \subfloat[Initial Pose Impact in KITTI-CVL Dataset]{
        \begin{minipage}[t]{0.49\textwidth}
            \includegraphics[width=0.495\textwidth]{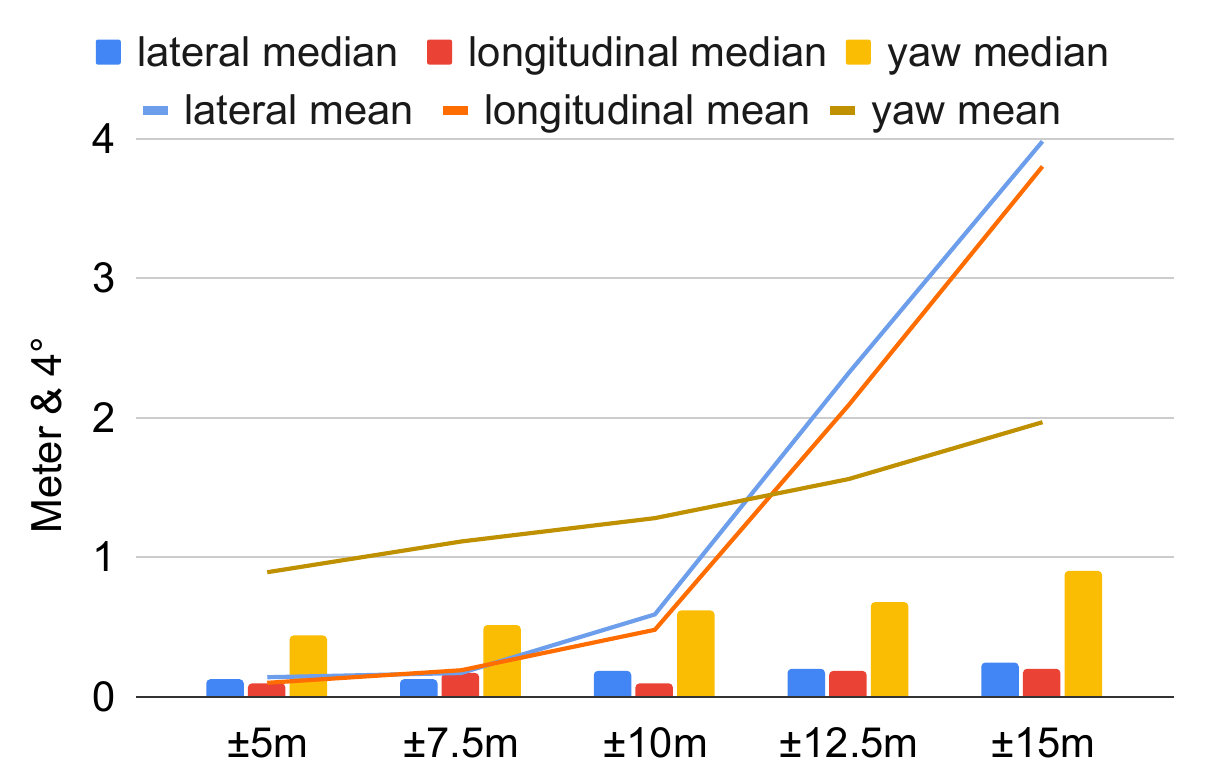}
            \includegraphics[width=0.495\textwidth]{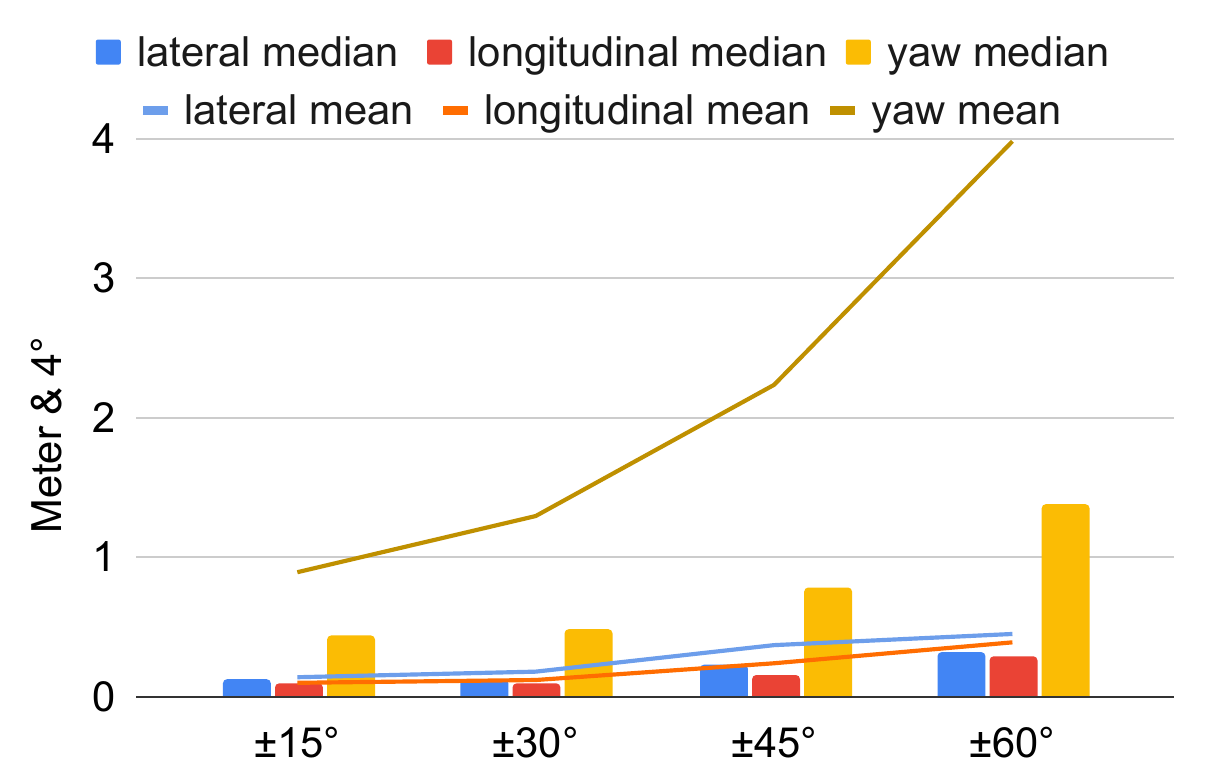}
        \end{minipage}
    }
    \\
    \subfloat[Initial Pose Impact in FordAV-CVL Dataset]{
        \begin{minipage}[t]{0.49\textwidth}
        \includegraphics[width=0.495\textwidth]{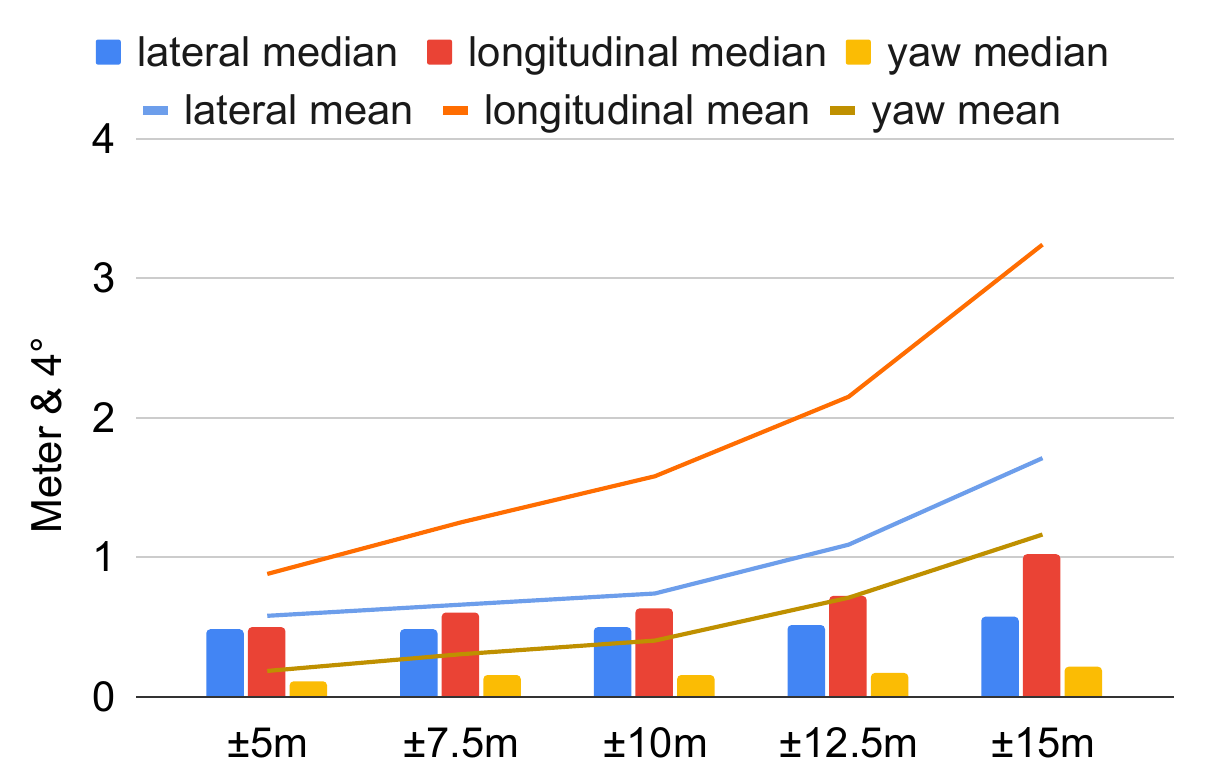}
        \includegraphics[width=0.495\textwidth]{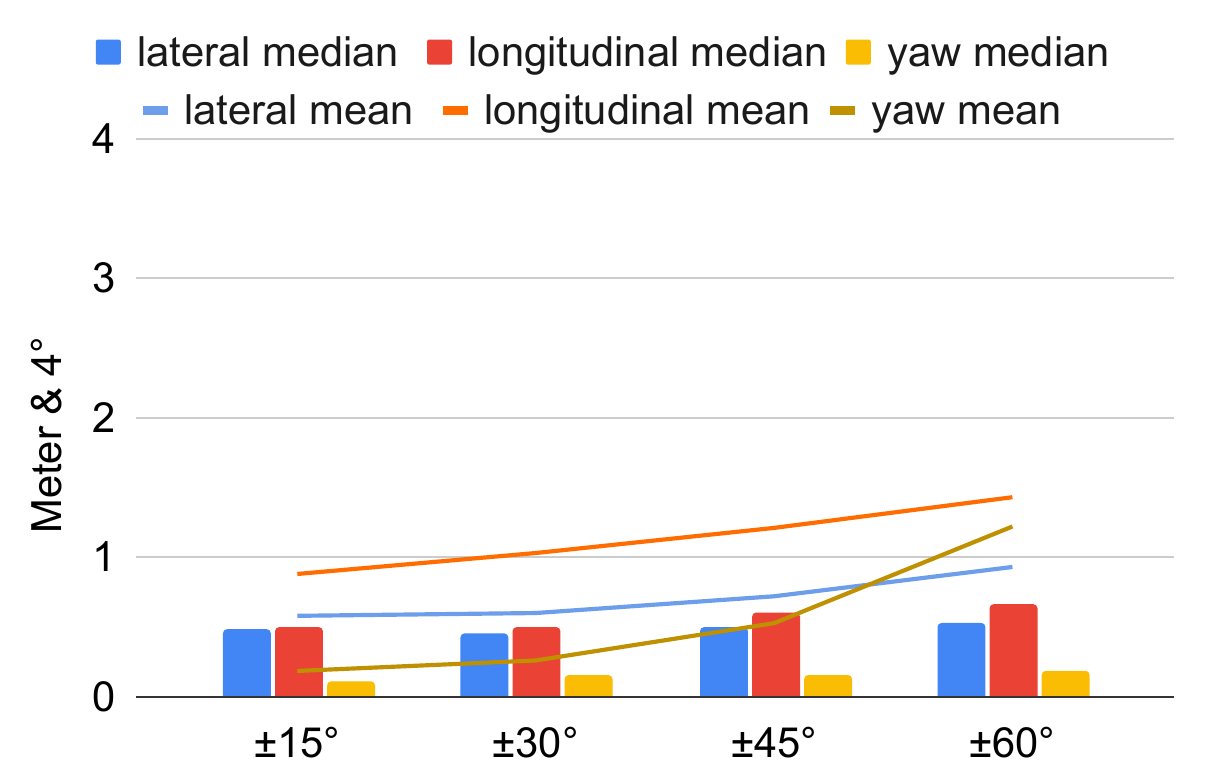}
        \end{minipage}
    }
     \caption{(left) Method performance as initial pose translation varies, with orientation noise fixed within $\pm15^\circ$ range. (right) Method performance as initial pose orientation varies, with translation noise fixed within $\pm5\mathrm{m}$ range. The vertical axis shows translation error in units of $\mathrm{m}$ and orientation error in units of 
     $4^{\circ}$. Additional metric results can be found in the supplementary.}
    \label{fig:converge}
    \vspace{-0.5cm}
\end{figure}

\noindent\textbf{Performance under Different Initial Poses}.
The proposed method utilizes the LM algorithm and is subject to a convergence range \footnote{The convergence range refers to the region in the pose space where the method can converge to the ground truth pose.} constraint. If the provided initial pose falls outside of this range, the method may fail to converge. 
To evaluate the method's robustness under a more stringent scenario, we conducted experiments using a comprehensive set of initial poses. The results, shown in Fig.~\ref{fig:converge}, indicate that our approach achieves a satisfactory level of accuracy even when the initial pose is subjected to yaw angle errors of up to $\pm60^\circ$ and lateral and longitudinal shifts of up to $\pm15$m. 
The longitudinal estimation is found to be more sensitive to the initial pose compared to the lateral estimation. Moreover, in KITTI-CVL datasets that rely solely on a front onboard camera, a larger difference between the mean and median values suggests more cases falling outside the convergence range. Therefore, the use of multiple camera inputs, such as in the FordAV-CVL dataset with four cameras, can significantly expand both the translation and orientation coverage ranges, with the orientation coverage range being notably more improved.

\section{Ablation Study}
\noindent\textbf{Two Confidence Maps}.
The proposed method adopts two types of confidence maps (\enquote{2c w/o SE}), i.e., view-consistent and on-ground maps.
An alternative approach was to use a single confidence map (\enquote{1c w/o SE}), which combined both on-ground and view-consistent confidences, and disabled gradient backpropagation from the satellite view.
A comparison of using different types of confidence maps is reported in Tab.~\ref{Tab:Ablation_study}. We can see that using two confidence maps with distinct gradient backpropagation mechanisms leads to better performance compared to the alternative approach.


\noindent\textbf{Spatial Embedding}.
We study the impact of Spatial Embedding by comparing the performance of our algorithm with (\enquote{Full}) and without Spatial Embedding (\enquote{2c w/o SE}), as shown in Tab.~\ref{Tab:Ablation_study}. The results demonstrate that incorporating Spatial Embedding significantly improves the performance of the PureACL algorithm.

\noindent\textbf{View-consistent On-ground Keypoint Detector}.
We compare our keypoint detection design with the SOTA SuperPoint \cite{DBLP:journals/corr/abs-1712-07629}. In this comparison, we use SuperPoint to detect keypoints and combine it with the two confidence maps to reduce the weights of points located on dynamic objects or above the ground plane. The results are presented in Tab.~\ref{Tab:Ablation_study} as \enquote{SuperPoint}. Our view-consistent on-ground point detector (\enquote{Full}) outperforms \enquote{SuperPoint} as it detects a sufficient number of on-ground keypoints, which is more beneficial for cross-view localization.

\begin{table}[!htb]
\renewcommand{\arraystretch}{1.0}
\caption{Ablation study on FordAV-CVL dataset}
\vspace{-0.1cm}
\label{Tab:Ablation_study}
\centering
\footnotesize 
\setlength\tabcolsep{1pt}
\begin{tabular}{c|c||@{\extracolsep{\fill}}c c|c c|c c} 
\hline
\multicolumn{2}{c||}{} & \multicolumn{2}{c|}{\bfseries Lateral} & \multicolumn{2}{c|}{\bfseries Longitudinal} & \multicolumn{2}{c}{\bfseries Yaw} \\
\multicolumn{2}{c||}{FordAV-CVL} & mean$\downarrow$ & median$\downarrow$ & mean$\downarrow$ & median$\downarrow$ & mean$\downarrow$ & median$\downarrow$ \\
\hline\hline
\bfseries 1c & \bfseries w/o SE & 0.63 & 0.49 & 1.29 & 0.68 & 2.08 & 1.28 \\
\bfseries 2c &\bfseries w/o SE & 0.63 & 0.48 & 1.17 & 0.63 & 0.90 & 0.57 \\
\multicolumn{2}{c||}{\bfseries Full} & \textbf{0.58} & \textbf{0.46} & \textbf{0.88} &\textbf{0.49} & \textbf{0.74} & \textbf{0.50}\\
\multicolumn{2}{c||}{\bfseries SuperPoint\cite{DBLP:journals/corr/abs-1712-07629}}& 0.65& 0.53 & 0.95 & 0.52 & 1.05 & 0.60\\
\multicolumn{2}{c||}{\bfseries Mean fusion} & 0.61 & 0.47 & 0.90 & 0.50 & 0.90 & 0.55 \\
\hline
\end{tabular}
\footnotesize
\begin{tablenotes}
\item[0] Our \textbf{Full} solution incorporates 2 confidence maps (2c) along with Spatial Embedding (w/ SE).
\vspace{-0.1cm}
\end{tablenotes}
\vspace{-0.1cm}
\end{table}

\noindent\textbf{Multi-camera Fusion Method}.
We compare two fusion methods for keypoints captured by multiple onboard cameras: selecting the highest-confidence 2D projection (\enquote{Full}), which is used in our proposed method, and computing the mean of features and confidence scores across all visible onboard camera images (\enquote{Mean fusion}). The results in Tab.~\ref{Tab:Ablation_study} show that highest-confidence fusion outperforms Mean fusion due to more reliable selection.


\section{Conclusion}
This paper presents PureACL, a novel cross-view localization approach for accurate 3-DoF pose estimation that supports flexible multi-camera inputs. Our approach utilizes a view-consistent on-ground keypoint detector to handle dynamic objects and viewpoint variations while removing off-the-ground objects to establish the homography transformer assumption. Additionally, PureACL incorporates a spatial embedding that maximizes the use of camera intrinsic and extrinsic information to reduce visual matching ambiguity. PureACL is the first sparse visual-only approach and the first visual-only cross-view method capable of achieving a mean translation error of less than one meter. Our future plan is to integrate PureACL into the SLAM system for reduced loop closure dependence. Ultimately, PureACL has the potential to lead to robust, reliable, accurate, and low-cost localization systems.

\section{Acknowledgements}
The research is funded in part by an ARC Discovery Grant (grant ID: DP220100800) to HL.

{\small
\bibliographystyle{ieee_fullname}
\bibliography{egbib}

\begin{thebibliography}{10}\itemsep=-1pt

\bibitem{Agarwal_2020}
Siddharth Agarwal, Ankit Vora, Gaurav Pandey, Wayne Williams, Helen Kourous,
  and James McBride.
\newblock Ford multi-{AV} seasonal dataset.
\newblock {\em The International Journal of Robotics Research},
  39(12):1367--1376, sep 2020.

\bibitem{barsan2020learning}
Ioan~Andrei Barsan, Shenlong Wang, Andrei Pokrovsky, and Raquel Urtasun.
\newblock Learning to localize using a lidar intensity map.
\newblock {\em arXiv preprint arXiv:2012.10902}, 2020.

\bibitem{DBLP:journals/corr/abs-1712-07629}
Daniel DeTone, Tomasz Malisiewicz, and Andrew Rabinovich.
\newblock Superpoint: Self-supervised interest point detection and description.
\newblock {\em CoRR}, abs/1712.07629, 2017.

\bibitem{engel2017direct}
Jakob Engel, Vladlen Koltun, and Daniel Cremers.
\newblock Direct sparse odometry.
\newblock {\em IEEE transactions on pattern analysis and machine intelligence},
  40(3):611--625, 2017.

\bibitem{feng2008gps}
Yanming Feng, Jinling Wang, et~al.
\newblock Gps rtk performance characteristics and analysis.
\newblock {\em Positioning}, 1(13), 2008.

\bibitem{fervers2022continuous}
Florian Fervers, Sebastian Bullinger, Christoph Bodensteiner, Michael Arens,
  and Rainer Stiefelhagen.
\newblock Continuous self-localization on aerial images using visual and lidar
  sensors.
\newblock In {\em 2022 IEEE/RSJ International Conference on Intelligent Robots
  and Systems (IROS)}, pages 7028--7035. IEEE, 2022.

\bibitem{geiger2013vision}
Andreas Geiger, Philip Lenz, Christoph Stiller, and Raquel Urtasun.
\newblock Vision meets robotics: The kitti dataset.
\newblock {\em The International Journal of Robotics Research},
  32(11):1231--1237, 2013.

\bibitem{google}
Google.
\newblock Maps static api, 2023.
\newblock 2023.

\bibitem{hampel2011robust}
Frank~R Hampel, Elvezio~M Ronchetti, Peter~J Rousseeuw, and Werner~A Stahel.
\newblock {\em Robust statistics: the approach based on influence functions},
  volume 196.
\newblock John Wiley \& Sons, 2011.

\bibitem{hu2018cvm}
Sixing Hu, Mengdan Feng, Rang~MH Nguyen, and Gim~Hee Lee.
\newblock Cvm-net: Cross-view matching network for image-based ground-to-aerial
  geo-localization.
\newblock In {\em Proceedings of the IEEE Conference on Computer Vision and
  Pattern Recognition}, pages 7258--7267, 2018.

\bibitem{kerl2013dense}
Christian Kerl, J{\"u}rgen Sturm, and Daniel Cremers.
\newblock Dense visual slam for rgb-d cameras.
\newblock In {\em 2013 IEEE/RSJ International Conference on Intelligent Robots
  and Systems}, pages 2100--2106. IEEE, 2013.

\bibitem{kingma2014adam}
Diederik~P Kingma and Jimmy Ba.
\newblock Adam: A method for stochastic optimization.
\newblock {\em arXiv preprint arXiv:1412.6980}, 2014.

\bibitem{langley1998rtk}
Richard~B Langley.
\newblock Rtk gps.
\newblock {\em Gps World}, 9(9):70--76, 1998.

\bibitem{Liu_2019_CVPR}
Liu Liu and Hongdong Li.
\newblock Lending orientation to neural networks for cross-view
  geo-localization.
\newblock In {\em The IEEE Conference on Computer Vision and Pattern
  Recognition (CVPR)}, June 2019.

\bibitem{liu2017efficient}
Liu Liu, Hongdong Li, and Yuchao Dai.
\newblock Efficient global 2d-3d matching for camera localization in a
  large-scale 3d map.
\newblock In {\em Proceedings of the IEEE International Conference on Computer
  Vision}, pages 2372--2381, 2017.

\bibitem{miller2021any}
Ian~D Miller, Anthony Cowley, Ravi Konkimalla, Shreyas~S Shivakumar, Ty Nguyen,
  Trey Smith, Camillo~Jose Taylor, and Vijay Kumar.
\newblock Any way you look at it: Semantic crossview localization and mapping
  with lidar.
\newblock {\em IEEE Robotics and Automation Letters}, 6(2):2397--2404, 2021.

\bibitem{mur2017orb}
Raul Mur-Artal and Juan~D Tard{\'o}s.
\newblock Orb-slam2: An open-source slam system for monocular, stereo, and
  rgb-d cameras.
\newblock {\em IEEE transactions on robotics}, 33(5):1255--1262, 2017.

\bibitem{nister2004visual}
David Nist{\'e}r, Oleg Naroditsky, and James Bergen.
\newblock Visual odometry.
\newblock In {\em Proceedings of the 2004 IEEE Computer Society Conference on
  Computer Vision and Pattern Recognition, 2004. CVPR 2004.}, volume~1, pages
  I--I. Ieee, 2004.

\bibitem{qian2019softtriple}
Qi Qian, Lei Shang, Baigui Sun, Juhua Hu, Hao Li, and Rong Jin.
\newblock Softtriple loss: Deep metric learning without triplet sampling.
\newblock In {\em Proceedings of the IEEE/CVF International Conference on
  Computer Vision}, pages 6450--6458, 2019.

\bibitem{Reid_2019}
Tyler~G.R. Reid, Sarah~E. Houts, Robert Cammarata, Graham Mills, Siddharth
  Agarwal, Ankit Vora, and Gaurav Pandey.
\newblock Localization requirements for autonomous vehicles.
\newblock {\em {SAE} International Journal of Connected and Automated
  Vehicles}, 2(3), sep 2019.

\bibitem{sarlin21pixloc}
Paul-Edouard Sarlin, Ajaykumar Unagar, Måns Larsson, Hugo Germain, Carl Toft,
  Viktor Larsson, Marc Pollefeys, Vincent Lepetit, Lars Hammarstrand, Fredrik
  Kahl, and Torsten Sattler.
\newblock {Back to the Feature}: Learning robust camera localization from
  pixels to pose.
\newblock In {\em CVPR}, 2021.

\bibitem{shi2020beyond}
Yujiao Shi and Hongdong Li.
\newblock Beyond cross-view image retrieval: Highly accurate vehicle
  localization using satellite image.
\newblock In {\em Proceedings of the IEEE Conference on Computer Vision and
  Pattern Recognition}, 2022.

\bibitem{shi2020looking}
Yujiao Shi, Xin Yu, Dylan Campbell, and Hongdong Li.
\newblock Where am i looking at? joint location and orientation estimation by
  cross-view matching.
\newblock In {\em Proceedings of the IEEE/CVF Conference on Computer Vision and
  Pattern Recognition}, pages 4064--4072, 2020.

\bibitem{shi2020optimal}
Yujiao Shi, Xin Yu, Liu Liu, Tong Zhang, and Hongdong Li.
\newblock Optimal feature transport for cross-view image geo-localization.
\newblock In {\em Proceedings of the AAAI Conference on Artificial
  Intelligence}, volume~34, pages 11990--11997, 2020.

\bibitem{shin2020dvl}
Young-Sik Shin, Yeong~Sang Park, and Ayoung Kim.
\newblock Dvl-slam: Sparse depth enhanced direct visual-lidar slam.
\newblock {\em Autonomous Robots}, 44(2):115--130, 2020.

\bibitem{tang2020rsl}
Tim~Yuqing Tang, Daniele De~Martini, Dan Barnes, and Paul Newman.
\newblock Rsl-net: Localising in satellite images from a radar on the ground.
\newblock {\em IEEE Robotics and Automation Letters}, 5(2):1087--1094, 2020.

\bibitem{tang2021get}
Tim~Y Tang, Daniele De~Martini, and Paul Newman.
\newblock Get to the point: Learning lidar place recognition and metric
  localisation using overhead imagery.
\newblock {\em Proceedings of Robotics: Science and Systems, 2021}, 2021.

\bibitem{tang2021self}
Tim~Y Tang, Daniele De~Martini, Shangzhe Wu, and Paul Newman.
\newblock Self-supervised learning for using overhead imagery as maps in
  outdoor range sensor localization.
\newblock {\em The International Journal of Robotics Research},
  40(12-14):1488--1509, 2021.

\bibitem{toker2021coming}
Aysim Toker, Qunjie Zhou, Maxim Maximov, and Laura Leal-Taix{\'e}.
\newblock Coming down to earth: Satellite-to-street view synthesis for
  geo-localization.
\newblock In {\em Proceedings of the IEEE/CVF Conference on Computer Vision and
  Pattern Recognition}, pages 6488--6497, 2021.

\bibitem{van2015world}
Frank Van~Diggelen and Per Enge.
\newblock The world’s first gps mooc and worldwide laboratory using
  smartphones.
\newblock In {\em Proceedings of the 28th international technical meeting of
  the satellite division of the institute of navigation (ION GNSS+ 2015)},
  pages 361--369, 2015.

\bibitem{von2020lm}
Lukas Von~Stumberg, Patrick Wenzel, Nan Yang, and Daniel Cremers.
\newblock Lm-reloc: Levenberg-marquardt based direct visual relocalization.
\newblock In {\em 2020 International Conference on 3D Vision (3DV)}, pages
  968--977. IEEE, 2020.

\bibitem{voraAerial}
Ankit Vora, Siddharth Agarwal, Gaurav Pandey, and James McBride.
\newblock Aerial imagery based lidar localization for autonomous vehicles,
  2020.

\bibitem{wang2023satellite}
Shan Wang, Yanhao Zhang, Ankit Vora, Akhil Perincherry, and Hengdong Li.
\newblock Satellite image based cross-view localization for autonomous vehicle.
\newblock In {\em 2023 IEEE International Conference on Robotics and Automation
  (ICRA)}, pages 3592--3599. IEEE, 2023.

\bibitem{wei2019learning}
Xinkai Wei, Ioan~Andrei B{\^a}rsan, Shenlong Wang, Julieta Martinez, and Raquel
  Urtasun.
\newblock Learning to localize through compressed binary maps.
\newblock In {\em Proceedings of the IEEE/CVF Conference on Computer Vision and
  Pattern Recognition}, pages 10316--10324, 2019.

\bibitem{6942558}
Ryan~W. Wolcott and Ryan~M. Eustice.
\newblock Visual localization within lidar maps for automated urban driving.
\newblock In {\em 2014 IEEE/RSJ International Conference on Intelligent Robots
  and Systems}, pages 176--183, 2014.

\bibitem{xia2022visual}
Zimin Xia, Olaf Booij, Marco Manfredi, and Julian~FP Kooij.
\newblock Visual cross-view metric localization with dense uncertainty
  estimates.
\newblock In {\em Computer Vision--ECCV 2022: 17th European Conference, Tel
  Aviv, Israel, October 23--27, 2022, Proceedings, Part XXXIX}, pages 90--106.
  Springer, 2022.

\bibitem{zhang2014loam}
Ji Zhang and Sanjiv Singh.
\newblock Loam: Lidar odometry and mapping in real-time.
\newblock In {\em Robotics: Science and Systems}, volume~2, pages 1--9.
  Berkeley, CA, 2014.

\bibitem{zhu2021vigor}
Sijie Zhu, Taojiannan Yang, and Chen Chen.
\newblock Vigor: Cross-view image geo-localization beyond one-to-one retrieval.
\newblock In {\em Proceedings of the IEEE/CVF Conference on Computer Vision and
  Pattern Recognition}, pages 3640--3649, 2021.

\end{thebibliography}
}

\clearpage
\onecolumn
\appendix
\etocdepthtag.toc{mtappendix}
\etocsettagdepth{mtchapter}{none}
\etocsettagdepth{mtappendix}{subsection}
\tableofcontents
\clearpage

\section{Evaluation of Other FordAV-CVL Dataset Logs}
The 'Log4' trajectory was chosen for method evaluation in SIBCL \cite{wang2023satellite} owing to its alignment accuracy with the satellite image. Furthermore, we also evaluated other logs from the FordAV-CVL Dataset in Tab.~\ref{Tab:FordAV_other} to supplement the results presented in Tab. 2 of the main paper. 
There are three travelings included in every log: '2017-08-04-26', '2017-07-24', and '2017-10-26'. For the purpose of training, evaluation, and test dataset split, we use '2017-07-24' as the evaluation dataset for all logs. We select the traveling sequence with a higher number of images as the training dataset. Specifically, the training dataset of 'Log1' and 'Log3' is '2017-10-26', whereas the training dataset of 'Log4' and 'Log5' is '2017-08-04-26'. 
The results demonstrate that our method is capable of estimating accurate 3-DoF pose with low spatial and angular errors in various scenarios, including freeway (log1), residential (log3, log5), university (log4) and vegetation (log4, log5).

\begin{table*}[!htb]
\renewcommand{\arraystretch}{1.0}
\caption{Performance of our method on additional logs of the FordAV-CVL dataset} 
\label{Tab:FordAV_other}
\centering
\small
\setlength\tabcolsep{1pt}
\begin{tabular*}{\textwidth}{c||@{\extracolsep{\fill}}c c c c c c|c c c c c c|c c c c c}
\hline
& \multicolumn{6}{c|}{\bfseries Lateral} & \multicolumn{6}{c|}{\bfseries Longitudinal} & \multicolumn{5}{c}{\bfseries Yaw} \\
& mean$\downarrow$ & median$\downarrow$ & 0.25m$\uparrow$ & 0.5m$\uparrow$ & 1m$\uparrow$ & 2m$\uparrow$ & mean$\downarrow$ & median$\downarrow$ & 0.25m$\uparrow$ & 0.5m$\uparrow$ & 1m$\uparrow$ & 2m$\uparrow$ &  mean$\downarrow$ & median$\downarrow$ & $1^\circ\uparrow$ & $2^\circ\uparrow$ & $4^\circ\uparrow$ \\
\hline\hline
Log1 & 0.64 & 0.37  & 34.98 & 62.45 & 84.39 & 91.94 & 0.25 & 0.12 & 82.86 & 90.40 & 95.07 & 98.47 & 2.37 & 0.70 & 58.84 & 70.84 & 81.23 \\
Log3 & 1.07 & 0.99  & 10.64 & 22.83 & 50.60 & 88.89  & 0.96  & 0.70 & 18.79 & 36.44 & 65.89 & 91.22 & 1.82 & 1.07  & 47.86 &  68.90 & 87.59\\
Log5 & 0.88 & 0.66  & 18.36 & 38.70 & 70.01 & 90.30 & 1.80 & 0.75 & 19.82 & 36.83 & 58.59 & 76.76 & 1.23 & 0.62  & 65.51 & 84.25 & 93.57 \\
ALL & 1.01 & 0.63  & 21.35 & 41.52 & 66.02 & 85.01 & 0.67 & 0.50  & 26.56 & 50.33 & 80.32 & 95.00 & 1.61 & 0.73  & 67.20 & 82.21 & 88.46\\
\hline
\end{tabular*}
\begin{tablenotes}
\item[0] ALL: All images from 'Log1', 'Log3', 'Log4', and 'Log5'. 'Log2' and 'Log6' are excluded due to the construction of road and building during data collection.
\end{tablenotes}
\vspace{-0.2cm}
\end{table*}

\section{Performance with Different Initial Poses}
In our main paper, we presented a chart illustration in Fig.7. Here, we further provide complete metrics results in Tab.~\ref{Tab:KITTI_initial} and Tab.~\ref{Tab:FordAV_initial}. 
By presenting these tables, we aim to provide a comprehensive view of the data and enable readers to analyze the metrics more thoroughly. 
\begin{table*}[!htb]
\renewcommand{\arraystretch}{1.0}
\caption{Performance of our method in different initial pose of the KITTI-CVL dataset} 
\label{Tab:KITTI_initial}
\centering
\small
\setlength\tabcolsep{1pt}
\begin{tabular*}{\textwidth}{c|c||@{\extracolsep{\fill}}c c c c c c|c c c c c c|c c c c c}
\hline
translation& yaw& \multicolumn{6}{c|}{\bfseries Lateral} & \multicolumn{6}{c|}{\bfseries Longitudinal} & \multicolumn{5}{c}{\bfseries Yaw} \\
m&$^\circ$& mean$\downarrow$ & median$\downarrow$ & 0.25m$\uparrow$ & 0.5m$\uparrow$ & 1m$\uparrow$ & 2m$\uparrow$ & mean$\downarrow$ & median$\downarrow$ & 0.25m$\uparrow$ & 0.5m$\uparrow$ & 1m$\uparrow$ & 2m$\uparrow$ & mean$\downarrow$ & median$\downarrow$ & $1^\circ\uparrow$ & $2^\circ\uparrow$ & $4^\circ\uparrow$ \\
\hline\hline
\multirow{4}{*}{15} & 7.5 & 0.17 & 0.12  & 84.28 & 98.44 & 99.42 & 99.55& 0.19 & 0.17 & 86.36 & 99.66 & 99.66 & 99.66 & 4.45 & 2.03  & 27.52 & 49.35 & 72.70 \\
 & 10 & 0.59 & 0.19 & 54.94 & 85.10 & 96.53 & 97.20  & 0.48  & 0.10 & 86.44 & 95.46 & 96.57 & 96.71 & 5.12 & 2.49  & 23.61 &  42.91 & 64.93\\
 & 12.5 & 2.32 & 0.20  & 58.54 & 79.51 & 83.57 & 84.72 & 2.10 & 0.18  & 70.68 & 81.65 & 81.71 & 81.78 & 6.24 & 2.70 & 23.07 & 41.34 & 60.64 \\
 & 15 & 3.98 & 0.25  & 50.25 & 64.27 & 67.24 & 69.28 & 3.80 & 0.20  & 58.45 & 64.63 & 64.81 & 65.06 & 7.87 & 3.59 & 19.89 & 35.32 & 52.96\\
 \hline
 30 & \multirow{3}{*}{5} & 0.18 & 0.13 & 76.99 & 96.40 & 99.89 & 99.95 & 0.12 & 0.10  & 94.76 & 99.64 & 99.88 & 99.91 & 5.18 & 1.92 & 30.86 & 51.04 & 69.59 \\
45 &  & 0.55 & 0.25 & 49.12 & 78.88 & 95.27 & 97.21 & 0.49 & 0.21  & 56.86 & 88.82 & 96.22 & 97.06 & 8.94 & 3.16 & 19.94 & 36.22 & 56.45 \\
60 &  & 0.45 & 0.33  & 39.73 & 70.11 & 95.94 & 99.26 & 0.39 & 0.29 & 44.14 & 79.55 & 97.18 & 99.12 & 16.04 & 5.48  & 15.47 & 28.21 & 43.88 \\
\hline
\end{tabular*}
\end{table*}

\begin{table*}[!htb]
\renewcommand{\arraystretch}{1.0}
\caption{Performance of our method in different initial pose of the FordAV-CVL dataset} 
\label{Tab:FordAV_initial}
\centering
\small
\setlength\tabcolsep{1pt}
\begin{tabular*}{\textwidth}{c|c||@{\extracolsep{\fill}}c c c c c c|c c c c c c|c c c c c}
\hline
translation& yaw& \multicolumn{6}{c|}{\bfseries Lateral} & \multicolumn{6}{c|}{\bfseries Longitudinal} & \multicolumn{5}{c}{\bfseries Yaw} \\
m&$^\circ$& mean$\downarrow$ & median$\downarrow$ & 0.25m$\uparrow$ & 0.5m$\uparrow$ & 1m$\uparrow$ & 2m$\uparrow$ & mean$\downarrow$ & median$\downarrow$ & 0.25m$\uparrow$ & 0.5m$\uparrow$ & 1m$\uparrow$ & 2m$\uparrow$ & mean$\downarrow$ & median$\downarrow$ & $1^\circ\uparrow$ & $2^\circ\uparrow$ & $4^\circ\uparrow$ \\
\hline\hline
\multirow{4}{*}{15} & 7.5 & 0.66 & 0.49  & 26.84 & 51.07 & 81.13 & 97.90 & 1.25 & 0.61  & 23.20 & 43.30 & 68.74 & 85.53 & 1.22 & 0.62  & 68.62 & 87.22 & 94.79 \\
 & 10 & 0.74 & 0.50  &26.40 & 50.01 & 79.61 & 96.58  & 1.58  & 0.64 & 22.22 & 41.55 & 65.80 & 81.96 & 1.61 & 0.64  & 66.92&  85.03 & 92.79\\
 & 12.5 & 1.09 & 0.53  & 25.57 & 47.92 &76.15 & 93.02 & 2.15 & 0.72  & 20.32 & 38.35 & 61.24 & 76.29 & 2.84 & 0.69 & 63.22 & 80.65 & 88.87 \\
 & 15 & 1.71 & 0.58  & 23.52 & 44.12 & 69.72 & 86.49 & 3.24 & 1.02  & 16.23 & 30.74 & 49.69 & 63.51 & 4.65 & 0.86 & 54.61 & 71.31 & 80.57\\
 \hline
 30 & \multirow{3}{*}{5} & 0.60 & 0.45  & 27.10 & 54.47 & 84.30 & 97.62 & 1.03 & 0.50  & 25.65 & 49.85 & 73.28 & 88.00 & 1.04 & 0.61  & 71.87 & 92.41 & 97.36 \\
45 &  & 0.72 & 0.50 & 26.72 & 49.96& 79.48 & 96.18  & 1.21  & 0.61 & 22.86 & 42.70 & 68.03 & 85.31 & 2.11 & 0.65  & 66.39 &  84.53 & 91.99\\
60 &  & 0.93 & 0.53  & 25.39 & 47.17 & 74.75 & 92.12 & 1.43 & 0.67  & 22.01 & 40.50 & 64.21 & 81.03 & 4.88 & 0.72  & 61.51 & 78.44 & 85.70 \\
\hline
\end{tabular*}
\end{table*}

\begin{figure}[!htb]
    \centering
    \includegraphics[width=0.49\textwidth]{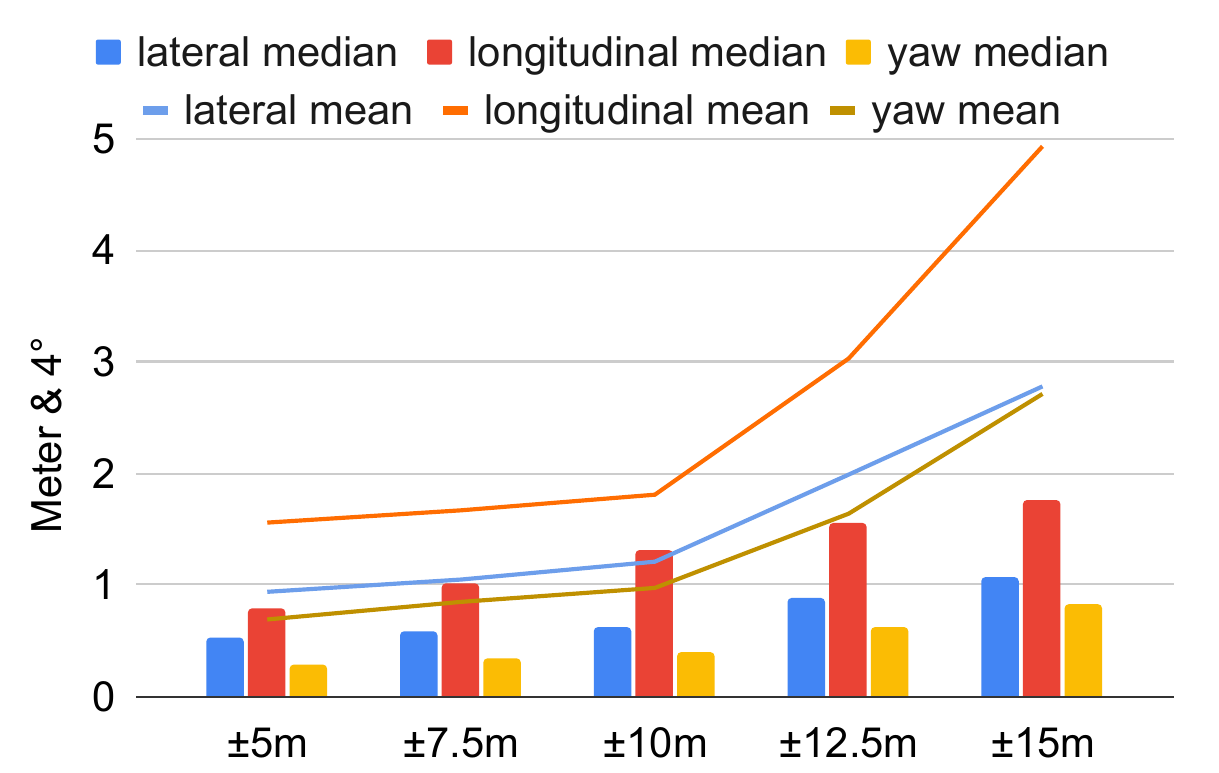}
    \includegraphics[width=0.49\textwidth]{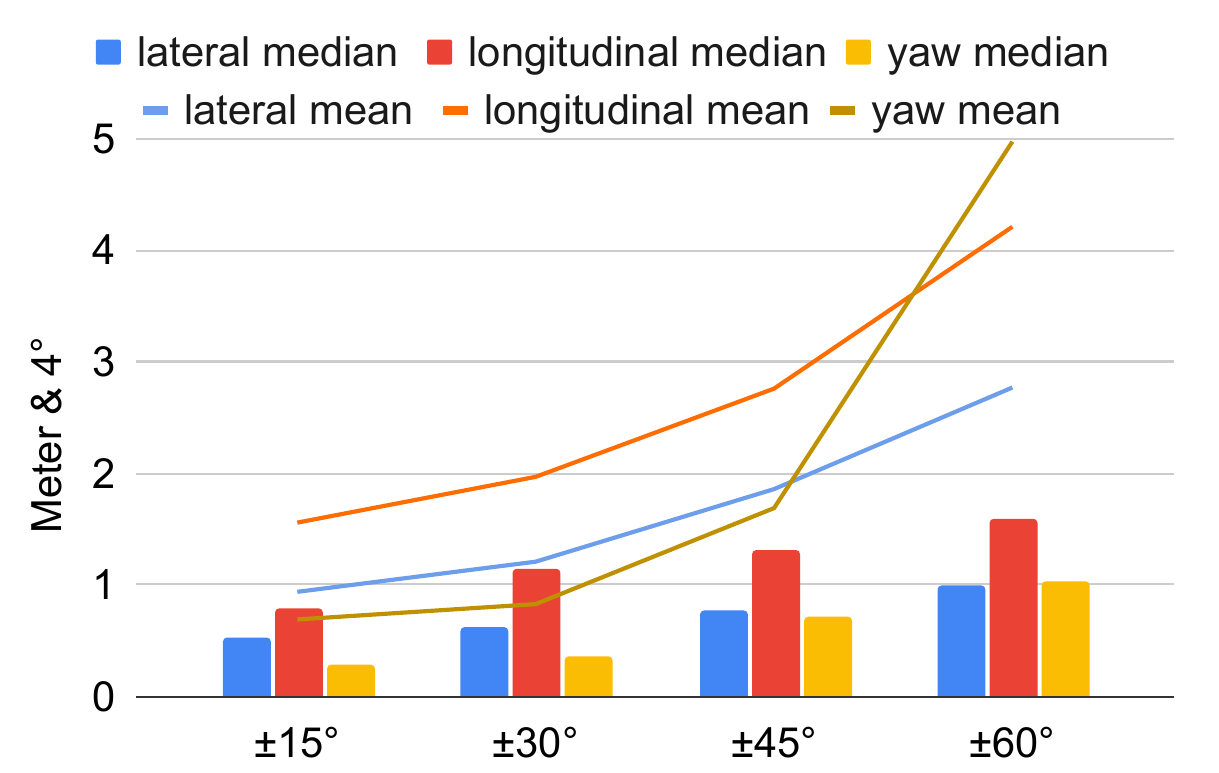}
    \caption{Impact of Initial Pose with one front camera setting in FordAV-CVL dataset. (left) Method performance as initial pose translation varies, with orientation noise fixed within $\pm15^\circ$ range. (right) Method performance as initial pose orientation varies, with translation noise fixed within $\pm5\mathrm{m}$ range.
    The vertical axis shows translation error in units of $\mathrm{m}$ and orientation error in units of $4^{\circ}$.}
    \label{fig:converge_ford_1c}
    \vspace{-0.2cm}
\end{figure}

In order to facilitate comparison, we have included a performance analysis with a single front onboard camera from the FordAV-CVL dataset, as depicted in Fig.~\ref{fig:converge_ford_1c}.

To bring a comprehensive view of the data and enable readers to analyze the metrics more thoroughly, we further provide complete metrics results in Tab.~\ref{Tab:KITTI_initial} and Tab.~\ref{Tab:FordAV_initial}, as a supplementary of the chat illustration in Fig.~7 in our main paper.
In order to facilitate comparison, we have included a performance analysis with a single front onboard camera from the FordAV-CVL dataset, as depicted in Fig.~\ref{fig:converge_ford_1c}.

\section{Visualization of Confidence Maps}
\begin{figure*}[!htb]
    \centering
        \includegraphics[width=0.99\textwidth]{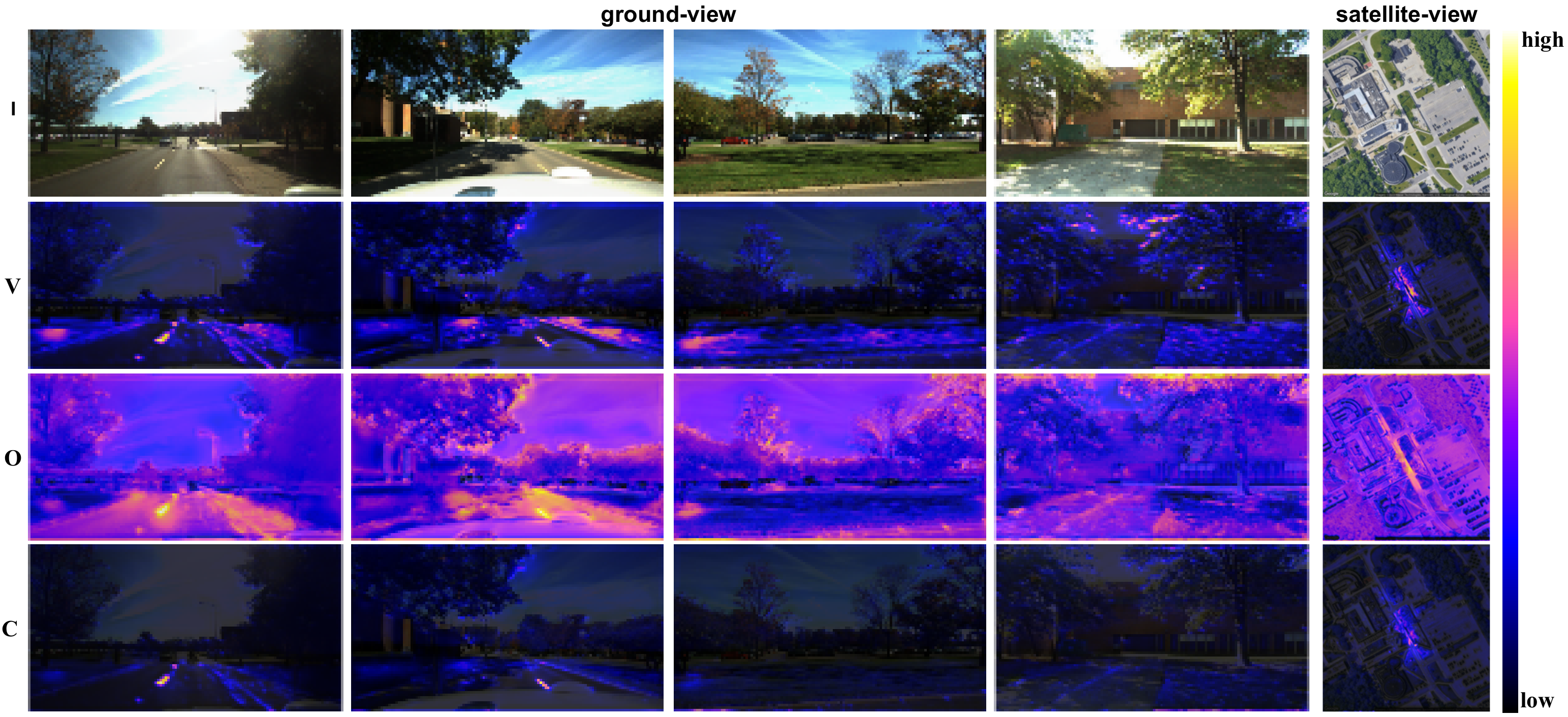}
    \caption{Illustration of confidence maps in FordAV-CVL dataset. The view-consistent confidence map ($2^{nd}$ row) $V$ assigns high confidence to objects that appear consistently in both ground-view and satellite images, such as road marks, and curbs. Conversely, the confidence map assigns low confidence to temporally inconsistent objects, such as vehicles and pedestrians. The on-ground confidence map ($3^{rd}$ row) $O$ highlights only on-ground cues. It is noteworthy that the on-ground confidence map $O$ assigns a high score to the sky and tree leaves. This is because the algorithm only uses key points located under the focal point of ground-view images. Consequently, objects that only exist in the upper part of ground-view images are not supervised and do not affect the localization. The fused confidence map ($4^{th}$ row) $C$ highlights objects that are both view-consistent and on-ground. 
    }
    \label{fig:confidence_ford}
\end{figure*}

The main paper provides an example of the view-consistent confidence map $V$, the on-ground confidence map $O$, and the fused confidence map $C$ of the KITTI-CVL dataset. Additionally, we present an example of the FordAV-CVL dataset in Fig.~\ref{fig:confidence_ford}. 
In addition, we have generated confidence map videos of continuous trajectories in both the KITTI-CVL and FordAV-CVL datasets, which can be found in the supplementary video.

\end{document}